\documentclass[preprint,5p]{elsarticle} %5p双栏

\usepackage{amssymb}
\usepackage{algorithm}
\usepackage{algpseudocode}
\usepackage{amsmath} % Include for use of \text in math mode
\usepackage[colorlinks=false, pdfborder={0 0 0}]{hyperref}
\usepackage{tabularx}
\usepackage{multirow}
\usepackage{booktabs} %三线表
\usepackage{caption}
\usepackage{graphicx}
\usepackage{natbib}

\usepackage{arydshln}    % 支持虚线
\usepackage{xcolor}      % 支持颜色

\usepackage[final]{changes}
\definecolor{deepgreen}{rgb}{0.0, 0.5, 0.0}
\definechangesauthor[color=red]{R1}
\definechangesauthor[color=deepgreen]{R2}
\definechangesauthor[color=blue]{R3}
\usepackage{todonotes}
\setcommentmarkup{\todo[color={authorcolor!20},size=\scriptsize]{#3: #1}}
  {\end{table}}
\usepackage{lineno}

\journal{ISPRS Journal of Photogrammetry and Remote Sensing}

\begin{document}
% \linenumbers % 开始编号
\begin{frontmatter}

  %% Title, authors and addresses

%% use the tnoteref command within \title for footnotes;
%% use the tnotetext command for theassociated footnote;
%% use the fnref command within \author or \address for footnotes;
%% use the fntext command for theassociated footnote;
%% use the corref command within \author for corresponding author footnotes;
%% use the cortext command for theassociated footnote;
%% use the ead command for the email address,
%% and the form \ead[url] for the home page:
%% \title{Title\tnoteref{label1}}
%% \tnotetext[label1]{}
%% \author{Name\corref{cor1}\fnref{label2}}
%% \ead{email address}
%% \ead[url]{home page}
%% \fntext[label2]{}
%% \cortext[cor1]{}
%% \fntext[label3]{}

\title{GLD-Road:A global-local decoding road network extraction model for remote sensing images}

\author[a,b]{Ligao Deng}
\author[a]{Yupeng Deng}

\author[a]{Yu Meng\corref{cor1}}
\ead{mengyu@aircas.ac.cn}
\cortext[cor1]{Corresponding author}

\author[a]{Jingbo Chen}
\author[a]{Zhihao Xi}
\author[a]{Diyou Liu}
\author[c]{Qifeng Chu}

\renewcommand{\thefootnote}{\alph{footnote}}

\affiliation[a]{organization={Aerospace Information Research Institute, Chinese Academy of Sciences},
            addressline={9 Dengzhuang South Road},%, Haidian District
            city={Beijing},
            postcode={101408}, 
            % state={},
            country={China}}
\affiliation[b]{organization={School of Electronic, Electrical and Communication Engineering, University of Chinese Academy of Sciences},
            addressline={1 East Yanqi Lake Road},
            city={Beijing},
            postcode={100049}, 
            % state={},
            country={China}}
\affiliation[c]{organization={Heilongjiang Geographic Information Engineering Institute},
            addressline={No. 2, Tiesan Street, Baojian Road, Nangang District},
            city={Harbin},
            postcode={150081},
            % state={},
            country={China}}

\begin{abstract}
    Road networks are essential information for map updates, autonomous driving, and disaster response. 
    However, manual annotation of road networks from remote sensing imagery is time-consuming and costly, whereas 
    deep learning methods have gained attention for their efficiency and precision in road extraction.
    Current deep learning approaches for road network extraction fall into three main categories: postprocessing methods based on semantic 
    segmentation results, global parallel methods and local iterative methods. Postprocessing methods introduce quantization errors, 
    leading to higher overall road network inaccuracies; global parallel methods achieve high extraction efficiency but 
    risk road node omissions; local iterative methods excel in node detection but have relatively lower extraction efficiency. 
    To address the above limitations, We propose a two-stage road extraction model with global-local decoding, named GLD-Road, which possesses the high efficiency of global parallel methods and the strong node perception capability of local iterative methods, enabling a significant reduction in inference time while maintaining high-precision road network extraction. In the 
    first stage, GLD-Road extracts the coordinates and direction descriptors of road nodes using global information from the 
    entire input image. Subsequently, it connects adjacent nodes using a self-designed graph network module (Connect Module) to 
    form the initial road network. In the second stage, based on the road endpoints contained in the initial road network, 
    GLD-Road iteratively searches local images and the local grid map of the primary network to repair broken roads, ultimately 
    producing a complete road network. Since the second stage only requires limited supplementary detection of locally missing nodes, GLD-Road significantly reduces the global iterative search range over the entire image, leading to a substantial reduction in retrieval time compared to local iterative methods. Finally, experimental results revealed that GLD-Road outperformed current state-of-the-art methods, 
    achieving improvements of 1.9\% and 0.67\% in average path length similarity (APLS) on the City-Scale and SpaceNet3 datasets, 
    respectively. Moreover, compared with those of a global parallel method (Sat2Graph) and a local iterative method (RNGDet++), 
    the retrieval time of GLD-Road exhibited reductions of 40\% and 92\%, respectively, suggesting that GLD-Road achieves a pronounced improvement in road network extraction efficiency compared to existing methods. The experimental results are available at https://github.com/ucas-dlg/GLD-Road.
\end{abstract}


\end{frontmatter}

%% main text
\section{Introduction}
As critical components of fundamental geographic information, road networks reflect the structures and spatial layouts of roads. They are typically stored in vector format, where vertices represent intersections and edges represent road segments\cite{haklay2008openstreetmap}. Road network extraction is essential for a wide range of applications, including map updating\cite{chen2024updating, bastani2021beyond}, autonomous driving\cite{liao2022maptr, yuan2024streammapnet}, disaster response\cite{ma2013automatic, huang2021damaged}, and urban planning\cite{chen2022road, qian2021rationality}. In these scenarios, precisely and efficiently extracting road networks is crucial, and the use of remote sensing imagery for road network extraction is an important method for achieving this goal. The traditional method of manually delineating road networks on remote sensing imagery is time-consuming and costly\cite{bastani2021beyond}. Deep learning demonstrates formidable capabilities in tasks such as image classification, image segmentation, and object detection within the field of computer vision\cite{krizhevsky2012imagenet, oquab2023dinov2, chen2018encoder, cheng2022masked, li2023mask, ren2016faster, he2018mask, carion2020end}, Therefore, in recent years, the automatic extraction of road networks from remote sensing imagery using deep learning also garners widespread attention due to its immense potential\cite{chen2022road, mena2003state, wang2016review, liu2024review}.

The common methods for extracting road networks from remote sensing imagery can be broadly divided into two categories. The first one binary road segmentation results through a semantic segmentation network, and this is followed by the use of complex postprocessing techniques such as morphological thinning\cite{zhang1984fast} to extract a road network from the skeletonized road segmentation results. Additionally, many studies have focused on improving the topological accuracy of semantic segmentation networks. For example, D-LinkNet\cite{zhou2018d} enhances the road extraction capability by employing dilated convolutions to increase the size of the receptive field. In addition, Mosinska et al.\cite{mosinska2018beyond} proposed a topological loss function that explicitly guides the utilized model to convergence during training, ensuring the accuracy of the obtained topological structure. DDCTNet\cite{gao2024ddctnet} leverages deformable spatial and dynamic channel-wise cross-transformer attention mechanisms to better capture the spatial details and channel features of roads, mitigating issues caused by road obstructions from trees and shadows. However, since these methods rely on pixel-level semantic segmentation results, segmentation networks tend to focus more on the prediction accuracy achieved for individual pixels rather than the completeness of the output road network topology, which often results in fragmented road networks.
\begin{figure}[htbp]
  \centering
  \includegraphics[width=\columnwidth]{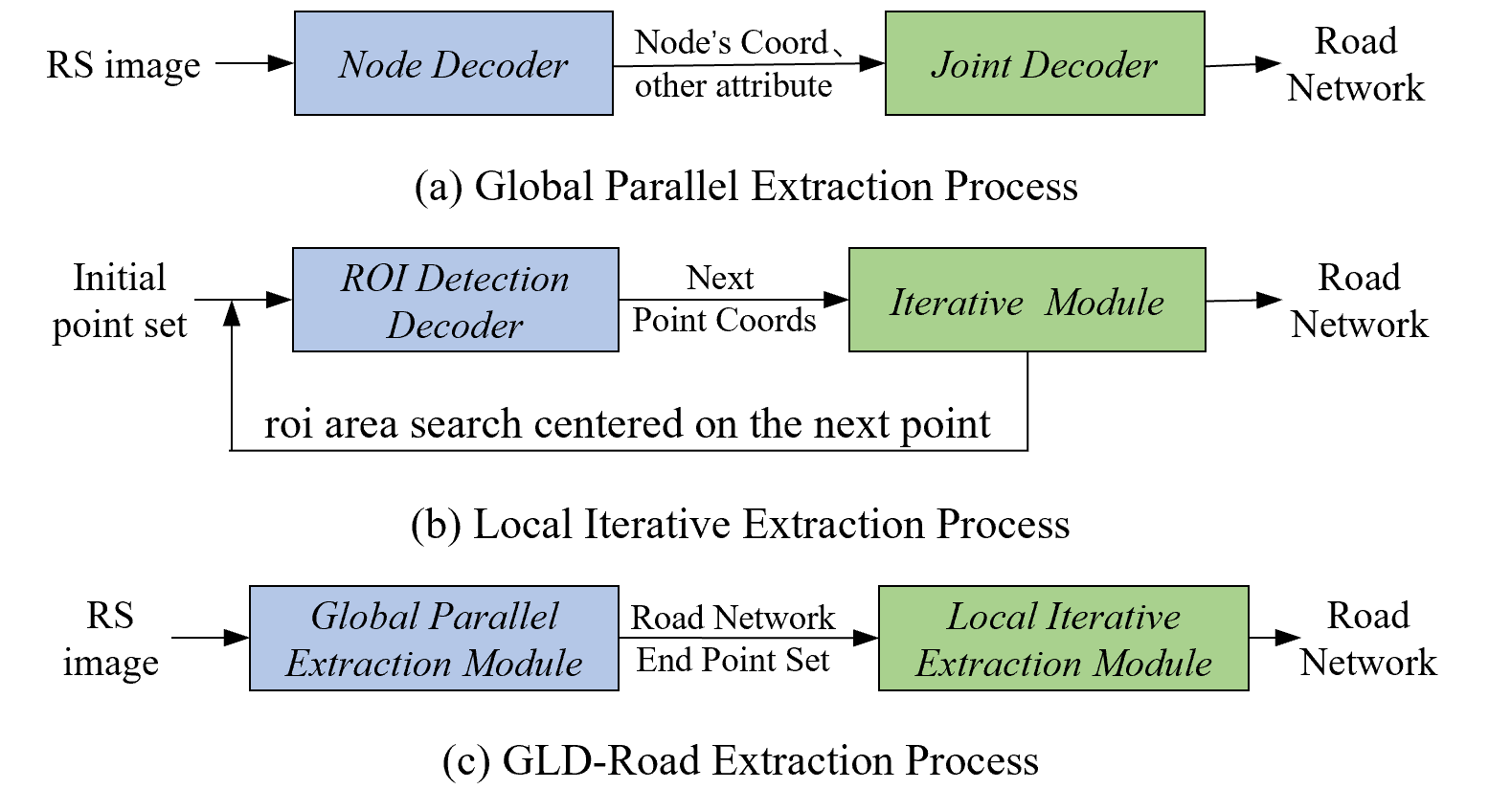} % 插入图片
  \caption{Comparison among the three types of road network extraction methods.}
  \label{fig:intro}
\end{figure}

\begin{figure}[htbp]
  \centering
  \includegraphics[width=\columnwidth]{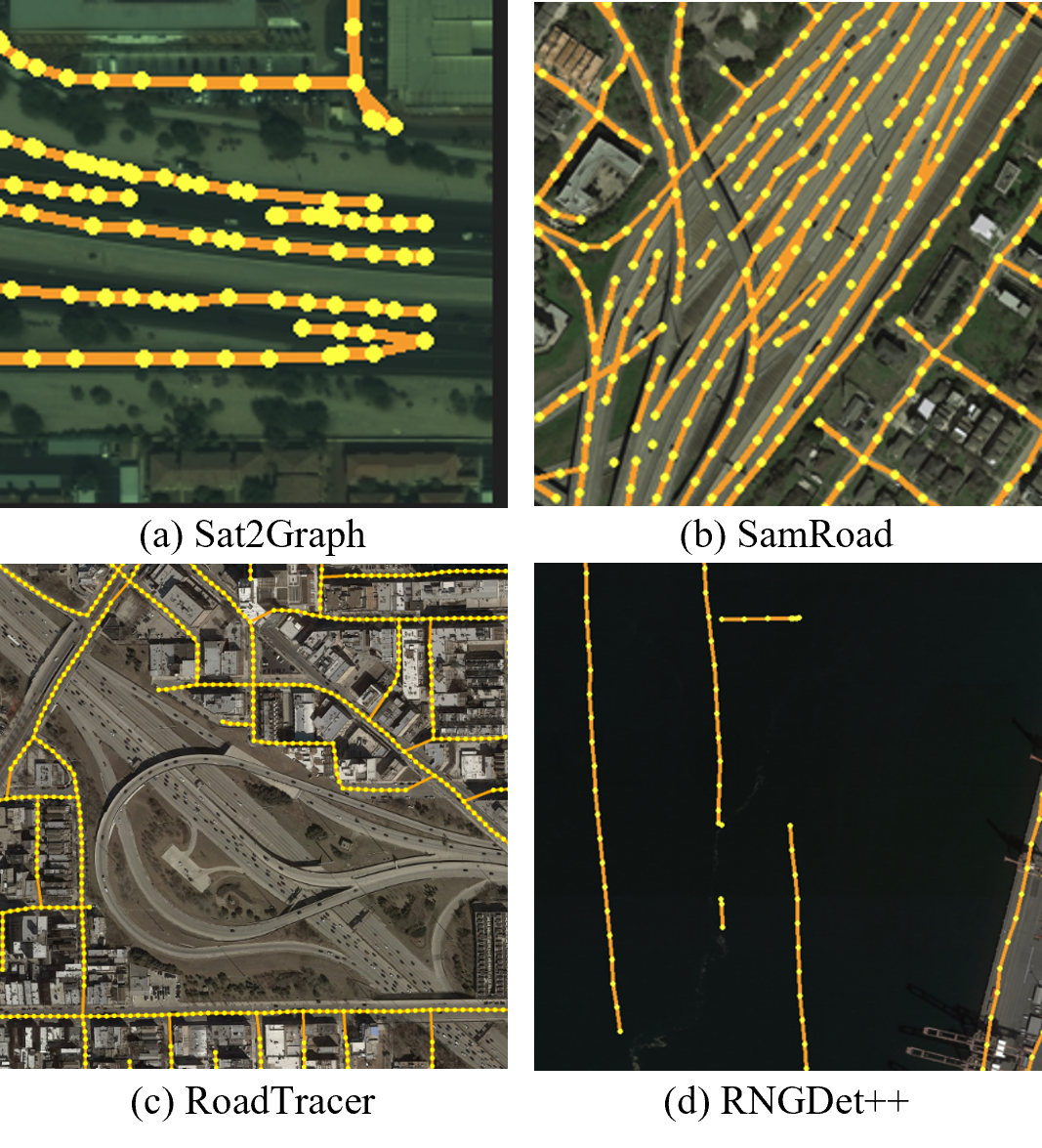} % 插入图片
  \caption{Visualization results of road network extraction using global parallel and local iterative methods.}
  \label{fig:fig2}
\end{figure}

The second one directly represents the road network as a graph structure. Specifically, roads are represented as an undirected graph G=(V,E), where V denotes the nodes within the road network and E represents the neighborhood connections between nodes. The construction of the road network graph structure can be refined into two methods : global parallel methods and local iterative methods. As shown in \autoref{fig:intro}(a), global parallel methods first extract the coordinates and attributes of road nodes, such as their directions, from the input image via a node decoder. The nodes are then connected via a node connection network or postprocessing algorithms to form a complete road network. Sat2Graph\cite{he2020sat2graph} uses tensor encoding to extract road networks, simultaneously obtaining road nodes and their directional information from images. TOPORoad\cite{zao2023topology} generates a road network by combining vertex connections with segmentation results. SamRoad\cite{10678570} first employs SAM\cite{kirillov2023segment} for feature extraction, and subsequently utilizes a node connection network to generate the final road network structure. These methods belong to the parallel category, first extracting all road nodes and then connecting them. Compared with iterative methods, parallel methods can more quickly extract road networks. \added[id=R3,comment={1}]{However, as shown in the first row of \autoref{fig:fig2}, typical global parallel methods (Sat2Graph\cite{he2020sat2graph} and SamRoad\cite{10678570}) exhibit an issue where road nodes fail to form effective connections. Since these methods predict each node independently and are influenced by interference from other regions in the global image, they often suffer from missing nodes or inaccurate connections, particularly in the middle of roads. Consequently, the resulting road network tends to be fragmented and lacks structural integrity.} In contrast, methods such as RoadTracer\cite{bastani2018roadtracer}, RNGDet\cite{xu2022rngdet}, and RNGDet++\cite{xu2023rngdet++} adopt iterative approaches to generate road networks. As shown in \autoref{fig:intro}(b), local iterative method randomly selects a point from the initial point set as the starting point. Beginning from this point, it uses the ROI Detection Decoder to identify successive nodes. The identified nodes are then passed to the Iterative Module, which continuously searches for and adds new nodes, progressively constructing a complete road network structure. Therefore, these methods are less efficient, with retrieval times often several times longer than those of global parallel methods. \added[id=R3, comment={1}]{Moreover, since the node retrieval process depends on the location of the initial or previous node, issues such as entire road segments being missed or error accumulation along long road stretches—as illustrated in the second row of \autoref{fig:fig2}—can occur.} Nevertheless, iterative methods offer notable advantages in maintaining road network connectivity. By searching for the next node within a local region, they leverage the position of the previous node and the existing road network to effectively reduce the impact of noise and enhance the structural consistency of subsequent node detection. This leads to improved connectivity and a reduction in fragmented road segments\cite{wang2023ternformer}.

To address issues such as node loss and disconnection in global parallel methods, as well as the slow retrieval speed in local iterative methods, we propose a two-stage road network-based extraction model built on a global-local strategy, termed GLD-Road. As shown in \autoref{fig:intro}(c), GLD-Road leverages the advantages of fast parallel processing and robust iterative node detection capabilities, thus it is divided into a global parallel extraction module and a local iterative extraction module. GLD-Road employs a unified model framework to merge the global parallel stage with the local iterative stage. In the first stage, GLD-Road processes the entire input image in parallel to extract the positions and orientations of all road nodes, connecting them via a self-designed graph network module (Connect Module) to form an initial road network. In the second stage, the model employs an iterative retrieval strategy centered on endpoints of the preliminary network, using local image and grid information to repair fragmented road segments, thereby completing the road network. \added[id=R3, comment={4}]{GLD-Road adopts a parallel approach to output road nodes in the first stage and connects the nodes through the Connect Module, resulting in a relatively short processing time. In the second stage, since only limited supplementary detection is required for locally missing areas, GLD-Road reduces the scope of global iterative search. Compared with other local iterative methods, it significantly shortens the retrieval time. Therefore, GLD-Road achieves not only high-precision road network extraction but also efficient and rapid road network extraction capabilities.}

Our main contributions are as follows:

1. This paper proposes a highly precise and efficient two-stage road network perception model, named GLD-Road, which is based on the core idea of combining global and local information. It includes a Global Query Decoder and a Local Query Decoder.

\sloppy
2. A denoising training strategy is proposed to alleviate the confusion in road node prediction. Additionally, a road node representation method based on points and 36-dimensional direction descriptors is introduced, providing more robust road node representations. Furthermore, a Connect Module is designed to avoid the difficult parameter adjustment process involved in traditional postprocessing-based connection algorithms.

3. The local grid features of the road network and the image features are fused, and an iterative retrieval method is used to re-examine the road endpoints in the initial road network, repair disconnected road nodes and improve the topological integrity of the road network.

\sloppy
4. We validate our approach on the City-Scale and \\ 
SpaceNet3 datasets. Experimental results indicate that, compared with other baseline methods, GLD-Road demonstrates superior performance in both road network extraction accuracy and efficiency on these datasets.

The rest of this paper is organized as follows: Section \hyperref[sec:section2]{2} reviews related work on road network extraction and object detection models. Section \hyperref[sec:section3]{3} presents a detailed explanation of the GLD-Road model design. Section \hyperref[sec:section4]{4} describes the experimental setup, comparative methods, and evaluation metrics. Section \hyperref[sec:section5]{5} provides an analysis and discussion of the experimental results. Finally, Section \hyperref[sec:section6]{6} concludes the paper.

\section{Related work}
\label{sec:section2}
\sloppy
In recent years, the extraction of road networks from remote sensing imagery has become a research hotspot\cite{chen2022road, liu2024review, kahraman2018road}. This paper discusses three categories of research methods that are closely related to our work: segmentation-based road network extraction methods, graph-based road network extraction methods, and object detection-related methods.

\subsection{Segmentation-Based Methods}
Most segmentation-based road network extraction methods typically involve two steps. First, a road segmentation network is used to extract road regions\cite{zhong2016fully, zhou2018unet++, chaurasia2017linknet, zhang2018road, chen2018encoder, zhou2018d, zhang2019fully, chen2022swin}, and then morphological thinning techniques\cite{zhang1984fast} are applied to the segmentation results to generate a single-pixel-width road network skeleton, which is further processed by postprocessing algorithms to connect and form the final road network. Zhang et al.\cite{zhang2018road} combined the advantages of ResNet\cite{he2016deep} and U-Net\cite{ronneberger2015u} to propose the Res-UNet network, which exhibits enhanced network depth and feature propagation capabilities, thus achieving promising results in road segmentation tasks. LinkNet\cite{chaurasia2017linknet} alleviates the information loss caused by encoder downsampling by connecting the features derived from the encoder and decoder. DlinkNet\cite{zhou2018d} integrates dilated convolution and LinkNet\cite{chaurasia2017linknet}, expanding the receptive field and improving the resulting road segmentation performance. Batra et al.\cite{batra2019improved} attained further enhanced road segmentation accuracy by jointly learning road masks, orientations, and segmentation results. Cheng et al.\cite{cheng2017automatic} proposed a cascaded convolutional neural network (CNN) that simultaneously extracts road and centerline probability maps, with the road centerlines refined by thinning techniques. DeepRoadMapper\cite{mattyus2017deeproadmapper} employs a shortest-path algorithm in its postprocessing stage to connect fragmented road networks. The region-based CNN (RCNN)-UNet\cite{yang2019road} adopts a multitask learning strategy that simultaneously detects roads and centerlines, with knowledge sharing implemented between the two tasks to achieve improved detection performance. BT-RoadNet was designed with a coarse map prediction module and a fine map prediction module, where the coarse module enhances road topology connections by introducing a spatial context module, and the fine module optimizes the boundaries obtained from the coarse results. DDCTNet\cite{gao2024ddctnet} utilizes a \replaced[id=R2,comment={A2}]{deformable and
dynamic cross-transformer}{DDCT} module and a \replaced[id=R2,comment={A2}]{cross-scale strippooling axial attention}{CSSA} structure to reduce road information losses and enhance linear road features, improving the accuracy of road extraction. However, due to the pixel-level semantic segmentation scheme used by these methods, their models fail to pay sufficient attention to global topological structures, and their results require complex postprocessing steps to form road centerlines, leading to lower topological correctness.

\subsection{Graph-Based Methods}
Graph-based methods can directly extract vectorized road networks from remote sensing imagery without the need for subsequent road thinning processes. RoadTracer\cite{bastani2018roadtracer} was the first model to adopt an iterative search method for road network detection; it constructs a decision function via a convolutional neural network and incrementally searches the entire input image by starting from a randomly selected road point. Owing to its use of fixed angles and step sizes, RoadTracer is prone to errors in complex intersection scenarios. RNGDet\cite{xu2022rngdet} and RNGDet++\cite{xu2023rngdet++} also employ an iterative search strategy, which uses the DETR network to detect the neighboring points of the current vertex and progressively generates a road network structure through iteration. If no neighboring points are found, the algorithm reverts to the previous node and continues the search process. Based on RNGDet++\cite{xu2023rngdet++}, DSVNet\cite{zhao2024deeply} introduces a deformable attention mechanism and designs a road vertex denoising training module to alleviate the confusion in vertex prediction, thereby improving road network extraction accuracy. Although these methods can directly generate road networks, their efficiency is relatively low because of their reliance on stepwise iterative searching. Additionally, since the node generation procedure depends on the previous node, error accumulation is likely. Sat2Graph\cite{he2020sat2graph} the encodes key points and directions within an image via 19-dimensional tensor encoding; this is followed by decoding and postprocessing steps, which generate a road network graph. However, due to the limitations of directional encoding, incorrect connections may occur. RelationFormer\cite{shit2022relationformer} improves upon the DETR model by detecting the relationships between objects while detecting the objects themselves and constructing connections between the road nodes. However, RelationFormer can only accurately handle small-scale images, and when stitching large-scale images, it is prone to the loss of topological integrity. TERNformer\cite{wang2023ternformer} introduces a \replaced[id=R2,comment={A2}]{depthwise separable dilated convolution blocks}{DSDB module} to extract more local features and an \replaced[id=R2,comment={A2}]{local structure exploring block}{LSED module} to enhance the topological structure of the constructed road network, acting as a topology-enhanced road network extraction method based on transformers. Although graph-based methods can directly obtain road network results, parallel and iterative strategies each have their own issues, such as missing road nodes and low retrieval efficiency. Combining the high retrieval efficiency of the parallel strategy with the strong node detection capability of the iterative strategy can further improve the accuracy of road network extraction.

\subsection{Transformer-Based Object Detection Methods}
Transformer\cite{vaswani2017attention} is a neural network model built on a self-attention mechanism, and it possesses advantages in capturing global contextual information and performing parallel computations. In recent years, transformers have been widely applied across various fields. DETR\cite{carion2020end} was the first model to adopt the transformer architecture for end-to-end object detection. DETR first extracts image features through a CNN and then sends the feature map and object queries to the transformer decoder, directly outputting the coordinates and classification results obtained for objects without the need to generate candidate boxes. However, DETR has shortcomings in terms of detecting small objects and its model convergence speed. Deformable DETR\cite{zhu2020deformable} introduces a deformable attention mechanism that focuses only on small-scale key points near the reference points, thereby achieving improved detection performance. \added[id=R2,comment={A3}]{DAB-DETR\cite{liu2022dabdetr} directly learns the four-dimensional coordinate anchor boxes as query, incorporating anchors to provide the model with positional priors and enhancing the interpretability of the query. Building upon this, DN-DETR\cite{li2022dn} addresses the instability in model training caused by Hungarian matching by introducing a novel approach. This method involves applying random flipping of labels, center shifting, and box scaling to the ground truth four-dimensional coordinates, bypassing Hungarian matching and directly computing the loss. This strategy effectively mitigates the instability issues associated with the matching process. Building upon the denoising training strategy of DN-DETR\cite{li2022dn}, DINO\cite{zhang2022dino} further introduces a contrastive denoising approach by generating positive and negative samples during the training process, addressing the issue of repeated outputs for the same object. Our GLD-Road adopts DINO's positive-negative sample denoising strategy during the training phase and refines it for the road network extraction task to enhance the accuracy of road node detection.}

\section{Methodology}
\label{sec:section3}
This chapter provides a detailed introduction to the components of the GLD-Road model. Section \hyperref[sec:section3_1]{3.1} introduces the overall architecture of GLD-Road. Section \hyperref[sec:section3_2]{3.2} introduces the Query Extractor module. Section \hyperref[sec:section3_3]{3.3} provides a detailed explanation of the Global Query Decoder module. Section \hyperref[sec:section3_4]{3.4} introduces the Local Query Decoder module. Section \hyperref[sec:section3_5]{3.5} introduces the denoising training module. Section \hyperref[sec:section3_6]{3.6} introduces the loss functions involved in each stage under the GLD-Road framework.

\subsection{Architecture Overview}
\label{sec:section3_1}
\begin{figure*}[ht]
  \centering
  \includegraphics[width=0.9\textwidth]{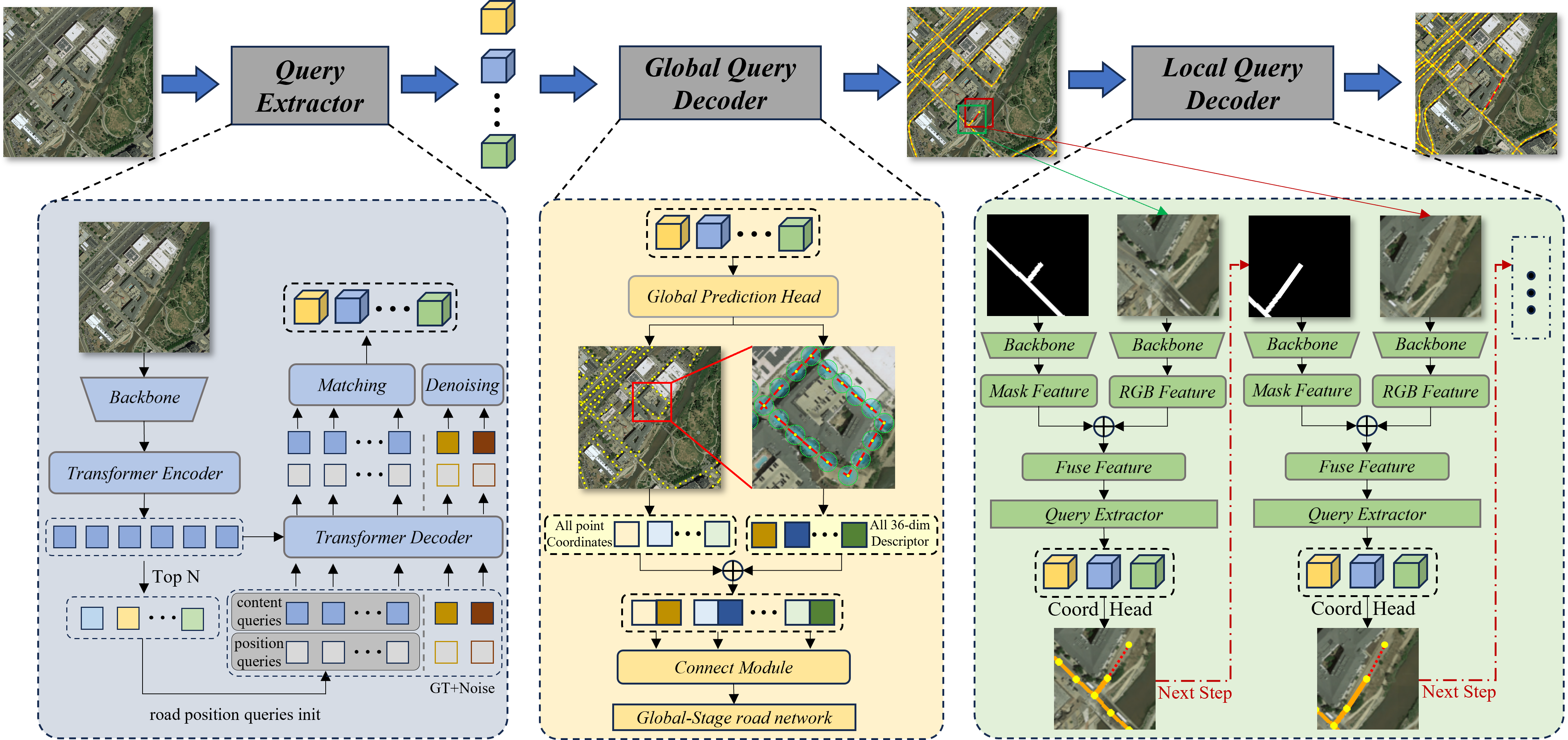}
  \caption{Structure of the GLD-Road model. Each square or cube represents a road query. The arrows indicate the direction of data flow. In the predicted road network result map, the orange lines represent the predicted road network, the yellow dots represent the nodes, and the red line segments indicate the iterative retrieval results derivedfrom the Local Query Decoder. In the \replaced[id=R2]{Global}{Globe} Query Decoder, the yellow dots indicate the positions of road nodes, while the red line segments represent the direction visualization results; the closer a line segment is to the circle's boundary, the higher the confidence in that direction.}\label{fig:over}
\end{figure*}

The overall structure of the GLD-Road model is shown in Figure~\ref{fig:over}. The model consists of three main components: Query Extractor, Global Query Decoder, and Local Query Decoder. The model takes an RGB remote sensing image as its input, and first, the Query Extractor module extracts all road queries from the image. These road queries are then processed through the Global Query Decoder, where the Global prediction head outputs the coordinates of the road nodes along with 36-dimensional directional descriptors. \added[id=R2, comment={A8}]{The node coordinates and the 36-dimensional directional descriptors are directly concatenated to form a 38-dimensional representation of each road node.} Next, the Connect Module models the connection relationships between the road nodes, generating a preliminary road network structure. In this preliminary road network, each road endpoint is treated as the center of a local image for the Local Query Decoder stage. In the Local Query Decoder module, the local grid results obtained for the preliminary road network and the corresponding local remote sensing image are processed through their respective backbones, generating Mask Features and RGB Features. These two types of features are concatenated to form Fuse Feature, which is then passed to the Query Extractor module to retrieve the adjacent subsequent nodes of the road endpoints. This process is repeated with each new node as the center until all endpoints are iteratively retrieved, filling in the gaps in the fragmented road network and producing a complete road network structure.

\subsection{Query Extractor Module}
\label{sec:section3_2}
The Query Extractor module is based on a multiscale deformable attention transformer architecture. Unlike traditional transformer architectures, this structure uses a deformable attention module to replace the self-attention and cross-attention modules. After the input image is processed by the Backbone, multiscale image features $f \in \mathbb{R}^{L_s \times 256}$ are obtained, where $ L_s = \sum_{k=2}^{5} \left(\frac{H}{2^k} \times \frac{W}{2^k}\right) $. These features are positionally encoded to maintain the positional relationships between patches and then fed into the Transformer Encoder. \added[id=R1,comment={6}]{The backbone extracts multi-scale preliminary features from the images, effectively reducing the computational complexity of the Transformer Encoder.} The Transformer Encoder consists of six layers of multi-scale deformable self-attention modules and feed-forward networks (FFNs). \added[id=R2,comment={4}]{The GLD-Road employs the multi-scale deformable attention (MSDA) module introduced by Zhu et al.\cite{zhu2020deformable}. In contrast to traditional self-attention mechanisms, MSDA sparsely samples a limited number of reference points across multi-scale features, significantly reducing computational overhead.} By enabling feature interaction across spatial positions, the Transformer Encoder effectively integrates global contextual information, thus enhancing the representation of long-range dependencies.

\added[id=R1, comment=4]{Integrating Conditional DETR with the road network extraction task, GLD-Road redefines the object queries in DETR as road queries, further categorizing them into road position queries and road content queries. These queries are responsible for encoding the positional and content features of road network nodes, respectively.} \added[id=R1, comment=5]{Drawing inspiration from the Two-Stage strategy in Deformable DETR and the initialization method for position queries in DINO, GLD-Road employs a three-layer multilayer perceptron (MLP) to filter encoder output features for initializing road position queries. Specifically, the MLP takes multi-scale image features from the encoder as input, with a shape of \([batch\_size, feat\_num, feat\_dims]\), and outputs confidence scores for each feature, shaped as \([batch\_size, feat\_num, score]\). Based on the MLP's output, the top \(N\) multi-scale features with the highest confidence scores are selected and processed by the Point Head to initialize road position queries. The detailed structure of the Point Head is described in Section \hyperref[sec:section3_3]{3.3}, and the value of \(N\) is determined by the complexity of the given dataset. Furthermore, road content queries are designed to be learnable according to the DINO framework.}

The Transformer decoder consists of six layers of multi-head attention modules, multi-scale deformable attention modules, and feedforward networks (FFN). To enhance the accuracy of road node prediction and accelerate model convergence, a denoising training module is incorporated into the Transformer decoder. During training, this module introduces noise to the ground-truth road node coordinates, generating positive and negative samples as additional decoder inputs. This approach enables the model to more effectively capture complex road structures. Further details on the denoising training module can be found in Section \hyperref[sec:section3_5]{3.5}.

The Transformer decoder takes as input the encoded feature representations, initialized road position queries, learnable road content queries, and the positive and negative road queries generated by the denoising module. Through the deformable attention mechanism, all road queries are iteratively updated across the decoder layers. The primary function of the Transformer decoder is to model the relationships among road queries and encode their contextual information. Finally, in the decoder output stage, bipartite matching is employed to associate road queries with ground-truth annotations and compute the corresponding loss. Since the denoising module explicitly identifies the ground-truth values corresponding to each positive and negative road query, no matching process is required, and the loss can be computed directly. This loss is referred to as the reconstruction loss in \autoref{fig:denoising}.
% Among the road queries output by the Transformer Encoder, the top N queries with the highest confidence levels are selected as the initialized road queries. The value of N depends on the complexity of the given dataset. The Transformer Decoder consists of 6 layers of multihead attention modules, multiscale deformable attention modules, and FFNs. Its input consists of the encoder output along with the initialized road queries. The Transformer Decoder is responsible for capturing the relationships between the queries and encoding contextual information into the queries. A bipartite matching technique is then applied to pair the queries with the ground-truth, after which the corresponding loss is computed. To improve the node prediction performance of the model and accelerate its convergence process, we introduce a denoising training module in the Transformer Decoder. During the training procedure of the decoder, this module adds noise to the ground-truth road node coordinates, generating both positive and negative samples as additional decoder inputs. This module enables the model to more effectively enhance its ability to detect complex road structures. The specific details of the denoising-based training module are described in Section \hyperref[sec:section3_5]{3.5}.

\subsection{Global Query Decoder}
\label{sec:section3_3}
The primary function of the Global Query Decoder is to connect the road node queries extracted by the Query Extractor through Global Prediction Head and Connect Module to generate an initial road network. Given the requirements for road node connections and road node modeling, the Global Prediction Head is divided into a Point Head and a Directional Head. The Head consists of three fully connected layers alternating with rectified linear unit (ReLU) activation functions. The difference between the two heads lies in the final output layer: the Point Head outputs the 2D coordinates (x, y) of the road nodes, whereas the \replaced[id=R2,comment={A7}]{Directional Head}{Direct Head} outputs a 36-dimensional directional descriptor.

\subsubsection{Road node modeling representation}
\begin{figure}[htbp]
  \centering
  \includegraphics[width=\columnwidth]{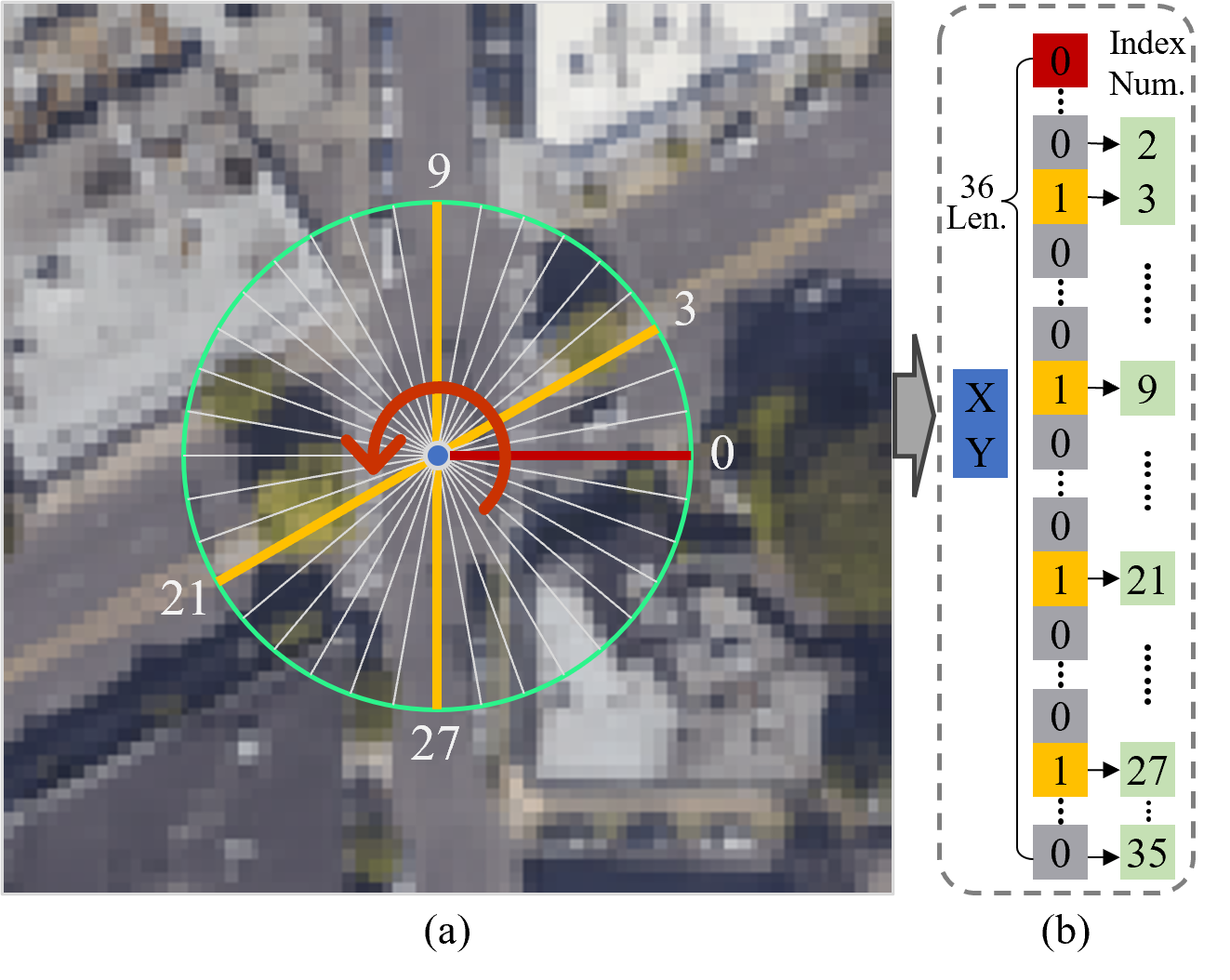} % 插入图片
  \caption{(a) Schematic diagram of road node modeling. (b) Node coordinates are shown on the left, the 36-dimensional directional descriptor vector (with zero values represented by ellipsis) is displayed in the middle, and the index numbers of the directional descriptor are indicated on the right.}
  \label{fig:road_node}
\end{figure}

Referring to the modeling methods of Sat2Graph\cite{he2020sat2graph} and TOPORoad\cite{zao2023topology}, road nodes are represented by road point coordinates and directional descriptors. However, these two methods suffer from significant quantization errors and cannot robustly represent node directions, particularly in scenarios with dense nodes, where incorrect road connections are prone to occur. To address this issue, we improve upon the original modeling methods by using a 36-dimensional directional descriptor to characterize the directional features of road nodes. As shown in \autoref{fig:road_node}, the road node modeling approach is as follows. The center of the circle represents the center point coordinates of the road, with the horizontal right direction being 0 degrees. A counterclockwise interval of 10 degrees is used for each direction. When the direction reaches 360 degrees, it coincides with the 0-degree direction, and the direction value is set to 0. As illustrated, for a certain road node, its neighboring road nodes exist in the 3rd, 9th, 21st, and 27th directions, and their directional representations are shown in \autoref{fig:road_node}(b). The coordinates are represented by a two-dimensional vector $(X,Y)$, and in the 36-dimensional directional descriptor vector, the 3rd, 9th, 21st, and 27th positions are marked as 1, while all other positions are set to 0.

% \begin{figure}[htbp]
%   \centering
%   \includegraphics[width=\columnwidth]{road_node.png} % 插入图片
%   \caption{(a) illustrates the road node modeling diagram, and (b) shows the node coordinates on the left and the directional descriptor vector on the right.}
%   \label{fig:road_node}
% \end{figure}

\subsubsection{Connect Module}
\added[id=R2, comment={A8}]{The input to the Connect Module is the 38-dimensional node feature representation formed by concatenating the predicted node coordinates with the 36-dimensional directional descriptors.} This module is a transformer-based modeling approach that determines whether a connection is present between each pair of predicted nodes within a local region of the image. Specifically, for a road node $P_v$, all neighboring nodes $\{P_n\}_{n=1}^{N_{pt}}$ within a given range $R$ are examined, and the Connect Module outputs the connection probability between $P_v$ and each of the $N_{pt}$ nodes $\{P_n\}_{n=1}^{N}$. If the probability exceeds the preset threshold, a connection is established between the two nodes.

\begin{figure}[htbp]
  \centering
  \includegraphics[width=\columnwidth]{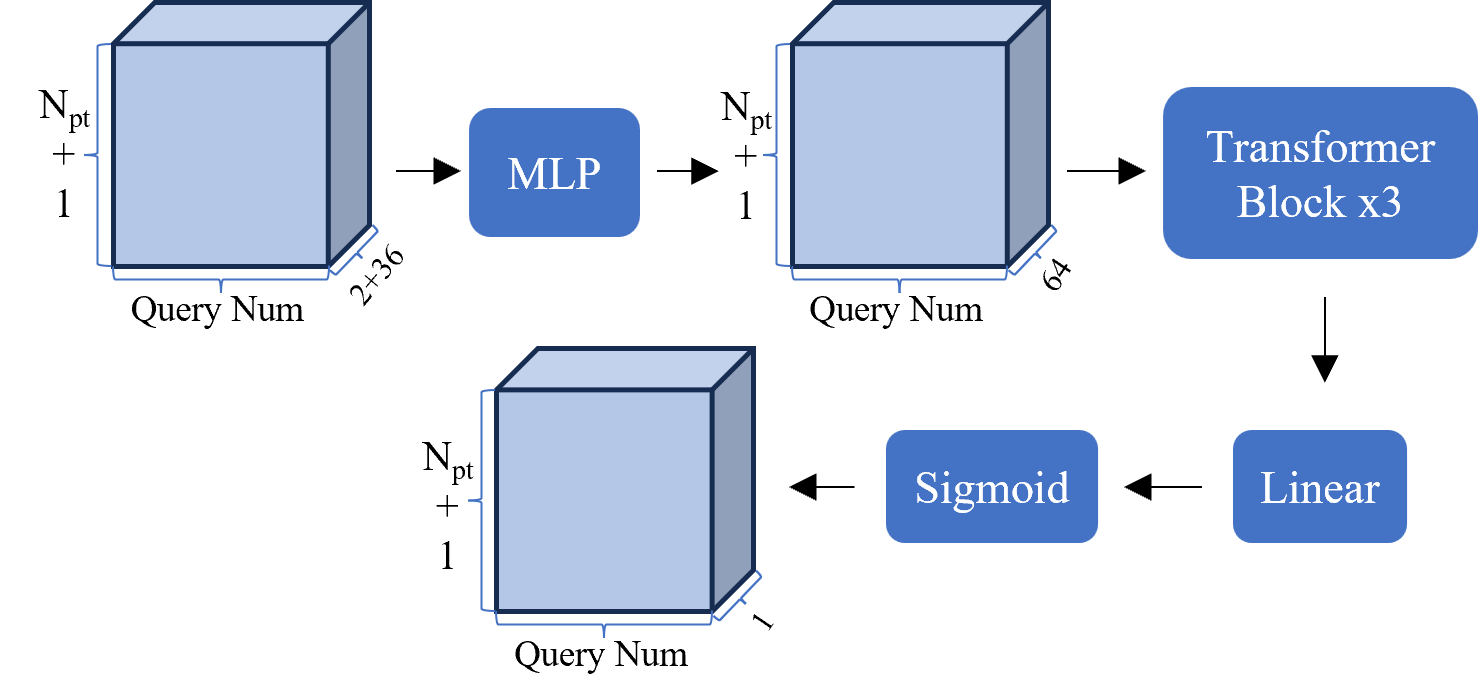} % 插入图片
  \caption{Structure of the Connect Module}
  \label{fig:connect_module}
\end{figure}

The Connect Module formulates the road node connection task as a probability prediction problem between nodes. The input of the Connect Module is a combination of node features and the corresponding neighboring node feature pairs $\left\{ \left( \text{Feat}^{P_v}, \text{Feat}^{P_n} \right) \mid 0 \leq n \leq N_{pt} \right\}$. As shown in \deleted[id=R2,comment={A3}]{Figure} \autoref{fig:connect_module}, these vectors are first projected to $ {(N_{pt}, 38)} $, which is followed by a ReLU activation function, and then projected again to feature vectors with sizes of $ {(N_{pt}, 64)} $. The feature vectors are then fed into a 3-layer multi-head self-attention module for feature interaction, with the feature dimensions remaining unchanged. Finally, the feature vectors are input into a fully connected layer and a sigmoid function produces a tensor with a size of $ {(N_{pt}, 2)} $, representing the connection probability between the nodes with values in the range $(0, 1)$.

\subsubsection{Connected Label Generation}
Since the predicted road nodes contain only directional and coordinate information, without connection relationships between the nodes, it is necessary to generate connection relationship labels during the training stage by mapping the predicted nodes to the ground-truth map. The label generation process consists of three steps: a) valid node filtering, b) mapping the predicted nodes to the ground-truth map, and c) generating connection relationships between the predicted nodes. Valid node filtering: First, the ground-truth road map is rasterized into line segments with a pixel width of 5. If a predicted node falls within the range of the line segment, the node is retained; otherwise, it is considered invalid and discarded; Mapping the predicted nodes to the ground-truth map: All valid nodes are traversed, and their Euclidean distances to the ground-truth centerline are calculated. The centerline points that are closest to each valid node are selected as the projection points of that valid node on the ground-truth map; Generating connection relationships between the predicted nodes: Since the roads in the ground-truth map are connected, the connection relationships between the projection points can be used to derive the corresponding connection relationships between the predicted nodes, thereby generating ground-truth labels for the connections between the predicted nodes.

\subsection{Local Query Decoder}
\label{sec:section3_4}
\added[id=R2, comment=3]{Before introducing the Local Query Decoder module, we first define the concepts of road endpoints. The initial road network is constructed by the previous Global Query Decoder stage and is represented as a graph \( G_{Global} = (V, E) \), where \( V \) denotes the set of nodes and \( E \) denotes the set of edges. We define \textit{road endpoints} as nodes with fewer than two adjacent nodes: $V_{end} = \left( v \in V \mid \text{len}(E) < 2 \right)$. 
Based on \( G_{Global} = (V, E) \), a road raster map \( M_{\text{road}} \) with a line width of 2 pixels is generated to represent the currently identified road network structure.}

\added[id=R2, comment=3]{Due to spectral differences among roads in the remote sensing image, the initial road network generated by the Global Query Decoder from global image features may contain disconnected or fragmented segments. To enhance the connectivity of the road network, we design a Local Query Decoder module that iteratively retrieves and completes broken road segments by leveraging local remote sensing imagery and the corresponding local region of \( M_{\text{road}} \) around road endpoints. The Local Query Decoder module primarily consists of the following four steps:}

\textbf{Step 1 Query Center Generation}:
We construct a set of query centers \( \{ v_k \}_{k=1}^{\text{Num}} \in V_{\text{end}} \) from the initial road network \( G_{\text{Global}} \). These endpoints are likely to indicate potential road extensions and are therefore selected as candidate starting points for local search.

\textbf{Step 2 Feature Extraction}:
A node \( v_k \) is randomly selected from the query center set \( \{ v_k \}_{k=1}^{\text{Num}} \), and a \( 128 \times 128 \) image patch centered at this point is cropped from both the remote sensing image and the road raster map \( M_{\text{road}} \). These two patches are then fed into two separate backbone networks with non-shared parameters to extract multi-scale spatial features. The extracted features are concatenated and subsequently passed to the Query Extractor module for further road node query extraction.

\added[id=R2, comment=3]{\textbf{Step 3 Iterative Node Generation}:
Based on the local road query, a Point Head module constructed using a three-layer MLP is employed to predict the positions (2D coordinates) of potential next road nodes. Since road structures consist of up to four connecting branches, each prediction may generate 0 to 4 nodes. The specific strategy is as follows:} 
\begin{itemize}
  \item If 0 nodes are generated: the query in the current region is considered unsuccessful. A new center is randomly selected from the candidate node set for the next query.
  \item If 1 node is generated: proceed to Step 4 to determine whether the node connects to the existing graph. If not, the node is treated as a new center and the query continues.
  \item If more than 1 node is generated: one node is randomly selected as the new center, and the remaining nodes are added to the candidate set for subsequent processing.
\end{itemize}

\added[id=R2, comment=3]{This process iterates until the candidate node set becomes empty, indicating that the final road network \( G_{\text{Final}} \) has been fully constructed.}

\textbf{Step 4 Check Node Connection}:
For each newly generated node, we determine whether it overlaps with an existing node in the graph \( G_{\text{Global}} \) or is within a distance of 2 pixels. If a connection is detected, the corresponding path segment is added to \( G_{\text{Global}} \); otherwise, the new node is treated as a new center and Steps 2 through 4 are repeated.
Through the above four steps, the Local Query Decoder effectively completes the disconnected segments in the initial road network and improves the overall connectivity of road extraction. The algorithm is detailed in Algorithm~\ref{alg:local_search}.

\begin{algorithm} \scriptsize
\caption{Local Query Decoder for Road Network Completion}\label{alg:local_search}
\begin{algorithmic}[1]
\State \textbf{Input:}
\State \quad Global-Stage primary road network $ G_{Global}=(V, E)$
\State \quad The endpoints $ V_{end}$
\State \quad An remote image $ I $
\State \textbf{Output:}
\State \quad The complete road network $ G_{Final}=(V, E) $
\State \While{$V_{end}$ is not empty}:
  \State $Step \gets 0$
  \State $v_k \gets V_{end}.\text{pop}() $
  \State $G_{Final} \gets G_{Global} $
  % \State $\text{roi_mask, roi_img} \gets \text{get_roi}(v_k, G, I)$
  % \State \texttt{roi\_mask, roi\_img} $\gets$ \texttt{get\_roi}$(v_k, G, I)$
  % \State $\text{roi\_mask, roi\_img} \gets \text{get\_roi}(v_k, G, I)$
  % \State $road\_nodes \gets \text{LocalQueryDecoder}(v_k, roi\_m, roi_img)$
  \State \While{$Step \leq 6$}:
    \State $Step \gets \text{Step + 1}$
    \State $roi_m, roi_{img} \gets \text{get\_roi}(v_k, G_{Final}, I)$
    \State $road\_nodes \gets \text{LocalQueryDecoder}(v_k, roi_m, roi_{img})$
    \If{$|road\_nodes| == 0$}
      \State \textbf{break}
    \ElsIf{$|road\_nodes| == 1$}
      \State Update $G_{Final}$
      \If{$\text{CheckNodeConnection}(road\_nodes, G_{Final})==\text{1}$}
        \State \textbf{break}
      \Else
        \State $v_k \gets road\_nodes$
      \EndIf
    \ElsIf{$|road\_nodes| > 1$}
      \State $v_k \gets \text{RandomSelectOne}(road\_nodes)$
      \State $V_{\text{end}} \gets V_{\text{end}} \cup (road\_nodes \setminus \{v_k\})$
      \State Update $G_{Final}$
      \State \textbf{break}
    \EndIf
  \EndWhile
\EndWhile
\State \Return $G_{Final}$
\end{algorithmic}
\end{algorithm}

\subsection{Denoising Training Strategy}
\label{sec:section3_5}
\begin{figure}[htbp]
  \centering
  \includegraphics[width=\columnwidth]{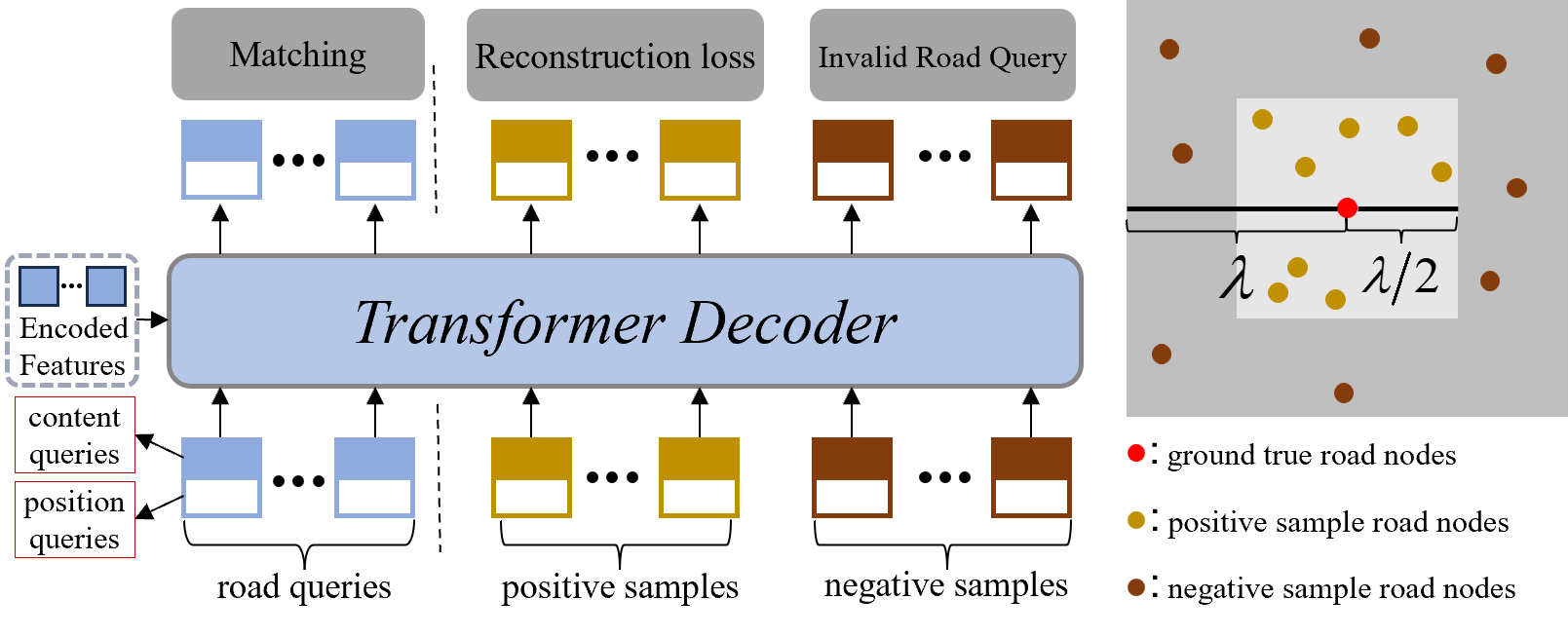} % 插入图片
  \caption{Structure of the denoising module. On the left, the denoising component of the Transformer Decoder is displayed, whereas on the right, the labels for positive and negative samples are visualized.}
  \label{fig:denoising}
\end{figure}

In densely populated road node regions, the dynamic matching process conducted during bipartite matching can lead to unstable model optimization results. During the inference stage, when multiple nodes are close to each other, prediction confusion may occur, reducing the topological accuracy of the road network. To address this issue, a denoising training module is additionally incorporated into the Query Decoder during the training phase, as shown in \autoref{fig:denoising}.

In the denoising training module, two random noises $\left( \Delta x_p, \Delta y_p \right)$ and $\left( \Delta x_n, \Delta y_n \right)$ are added to the coordinates of the ground-truth nodes, where the noise range is defined as $\left\{ \left| \Delta x_p \right|, \left| \Delta y_p \right| \right\} \leq \frac{\lambda}{2}$ for positive samples and $\frac{\lambda}{2} < \left\{ \left| \Delta x_p \right|, \left| \Delta y_p \right| \right\} \leq \lambda$ for negative samples. Here, $\lambda$ is a hyperparameter representing the noise magnitude, and in GLD-Road, it is set to 10. \added[id=R2,comment={A10}]{This means that the coordinates of positive samples are perturbed within the range of [-5, 5] pixels from the ground-truth coordinates, while the coordinates of negative samples fluctuate within the ranges of [-10, -5) $\cup$ (5, 10] pixels.} \added[id=R1,comment={1}]{The ground-truth nodes need to be processed into two types of queries: the road position queries and the road content queries. Specifically, we employ a learnable embedding layer to transform the ground-truth labels into a continuous 128-dimensional embedding space, which constitutes the road content queries. For the road position queries, we introduce noise perturbations to the ground-truth coordinates, normalize the perturbed coordinates, and then apply the Inverse Sigmoid function to ensure numerical stability, ultimately forming a 2D anchor.} \added[id=R1, comment=2]{Notably, during the model training process, the input of the Query Decoder in the Transformer Decoder includes the encoded features, the initialized road position queries, the learnable road content queries, as well as the positive and negative samples.} \added[id=R1,comment={3}]{However, During inference, the Transformer Decoder does not require positive or negative samples from the denoising component as input. In the training process, since the ground truth for positive and negative samples is known, bidirectional matching is not necessary within the denoising module. The introduction of the denoising module effectively mitigates prediction confusion among road nodes, thereby improving the topological accuracy of the road network.}

\added[id=R1,comment={7}]{We explicitly highlight the differences between our denoising strategy and those of DN-DETR and DINO. Our denoising strategy introduces both positive and negative samples for contrastive denoising, whereas DN-DETR generates only one type of noisy sample. It adopts a 2D xy anchor for position query, while DN-DETR uses a 4D xywh anchor. Noise sample generation is controlled by a single hyperparameter, unlike DN-DETR, which requires two for center shifting and box scaling. Since our task involves only a single class, label noise is omitted, unlike in DN-DETR. Compared to DINO, our method remains the same in contrastive denoising but differs in the other aspects.}

\subsection{Model Loss function}
\label{sec:section3_6}
The model is trained in two stages, resulting in two types of loss functions. These functions are referred to as global stage and the local stage loss functions in this section. \added[id=R1, comment=1]{Additionally, a reconstruction loss function is introduced in the denoising training strategy.} Below, each of the three loss functions is introduced separately.
\subsubsection{Global-Stage Loss Function}
The loss function in this stage is composed of four parts: \replaced[id=R2,comment={A12}]{$\mathcal{L}_{g-coord}$}{$L_coord$} for the road node position loss, \replaced[id=R2,comment={A12}]{$\mathcal{L}_{direct}$}{$L_direct$} for the road node direction loss, \replaced[id=R2,comment={A12}]{$\mathcal{L}_{coonect}$}{$L_coonect$} for the road network structural connectivity loss, \added[id=R1, comment=1]{and $\mathcal{L}_{g-reconstruction}$ for the reconstruction loss of denoised samples.}
% \added[id=R2]{
%   \begin{align}
%     L_{{global}} =  \nonumber \\
%     \lambda_{{g-coord}} L_{{g-coord}} + \lambda_{\text{direct}} L_{\text{direct}} + \lambda_{\text{connect}} L_{\text{connect}}
%     % &L_{{global}} = \lambda_{{g-coord}} L_{{g-coord}} + \lambda_{\text{direct}} L_{\text{direct}} \nonumber \\
%     % &+ \lambda_{\text{connect}} L_{\text{connect}}
%   \end{align}
% }
% \begin{align}
%   L_{{global}} =  \nonumber \\
%   \lambda_{{g-coord}} L_{{g-coord}} + \lambda_{\text{direct}} L_{\text{direct}} + \lambda_{\text{connect}} L_{\text{connect}}
%   % &L_{{global}} = \lambda_{{g-coord}} L_{{g-coord}} + \lambda_{\text{direct}} L_{\text{direct}} \nonumber \\
%   % &+ \lambda_{\text{connect}} L_{\text{connect}}
% \end{align}
\added[id=R2, comment={A12}]{
  \begin{equation}
    \begin{split}
    \mathcal{L}_{\mathrm{global}} = \lambda_{\mathrm{g-coord}} \mathcal{L}_{\mathrm{g-coord}} + \lambda_{\mathrm{direct}} \mathcal{L}_{\mathrm{direct}} \\
    + \lambda_{\mathrm{connect}} \mathcal{L}_{\mathrm{connect}} + \lambda_{\mathrm{g-reconstruction}} \mathcal{L}_{\mathrm{g-reconstruction}}
    \end{split}
    \label{eq:loss_global}
  \end{equation}
}

\replaced[id=R2, comment={A11}]{$\lambda_{g-coord}$, $\lambda_{direct}$, $\lambda_{connect}$, and $\lambda_{g-reconstruction}$}{$\lambda_1$, $\lambda_2$, and $\lambda_3$} are the coefficients used to balance the loss terms. Since the positions of road nodes are unstable during the early stages of the model training process, the model focuses primarily on the road node position loss at the beginning. As the model converges, the weights for the direction loss and road node connection loss increase exponentially. Specifically, \replaced[id=R2, comment={A11}]{$\lambda_{g-coord}$}{$\lambda_1$} is set to 2, \added[id=R1, comment=1]{$\lambda_{g-reconstruction}$ is set to 1}, \replaced[id=R2, comment={A11}]{$\lambda_{direct}$}{ $\lambda_2$} is set to $ 2 \times e^{epoch-100}$ and \replaced[id=R2, comment={A11}]{$\lambda_{connect}$}{$\lambda_3$} is set to $ 5 \times e^{epoch-100}$, where $epoch$ represents the current training epoch.

$\mathcal{L}_{g-coord}$: The road node coordinate loss is based on the L1 loss. $P_{g-coord}$ represents the predicted xy coordinates of the road nodes, and $Y_{g-coord}$ represents the ground-truth xy coordinates of the road nodes:
\begin{equation}
  \added[id=R2]{\mathcal{L}_{g-coord} = L1(P_{g-coord}, Y_{g-coord})}
\end{equation}
$\mathcal{L}_{direct}$: Since most road nodes have only two directions in the road network and nodes with three or more directions occur mainly at intersections, to handle the class imbalance problem, the road node direction loss is computed with the focal loss. $P_{direct}$ represents the predicted road node direction, and $Y_{direct}$ represents the ground-truth road node direction:
\begin{equation}
  \mathcal{L}_{direct} = FocalLoss(P_{direct}, Y_{direct})
\end{equation}
$\mathcal{L}_{connect}$: For the road connectivity part, each node connection is formulated as a binary classification problem between the predicted and true connections. The binary cross entropy loss with logits (BCEWithLogitsLoss) is used to calculate this loss, where $P_{connect}$ epresents the predicted node connection and $Y_{connect}$ represents the ground-truth node connection:
\begin{equation}
  \mathcal{L}_{connect} = BCEWithLogitsLoss(P_{connect}, Y_{connect})
\end{equation}

\subsubsection{Local-Stage Loss Function}
The loss function in this stage is composed of three parts: \replaced[id=R2, comment={A12}]{$\mathcal{L}_{l-coord}$}{$L_coord$}, which represents the loss function for the road node coordinates, \replaced[id=R2, comment={A12}]{$\mathcal{L}_{prob}$}{$L_prob$}, which represents the effective probability of the predicted points, and \added[id=R1, comment=1]{$\mathcal{L}_{l-reconstruction}$, which represents the reconstruction loss of denoised samples.}

\added[id=R2, comment={A12}]{
  \begin{equation}
    \begin{split}
      \mathcal{L}_{local} = \lambda_{l-coord} \mathcal{L}_{l-coord} + \lambda_{prob} \mathcal{L}_{prob} \\
      + \lambda_{{l-reconstruction}} \mathcal{L}_{{l-reconstruction} }
    \end{split}
    \label{eq:loss_local}
  \end{equation}
}
% \begin{equation}
%   \added[id=R2]{\mathcal{L}_{local} = \lambda_{l-coord} \mathcal{L}_{l-coord} + \lambda_{prob} \mathcal{L}_{prob} \\
%    + \lambda_{\mathrm{g-reconstruction}} \mathcal{L}_{\mathrm{g-reconstruction} }}
% \end{equation}
where \replaced[id=R2, comment={A11}]{$\lambda_{l-coord}$}{$\lambda_4$}, \replaced[id=R2, comment={A11}]{$\lambda_{prob}$}{$\lambda_5$}, and $\lambda_{l-reconstruction}$ are the coefficients for balancing the loss terms; they are set to 2, 5 and 1, respectively, on the basis of empirical evidence.

The coordinate loss \replaced[id=R2, comment={A12}]{$\mathcal{L}_{coord}$}{$L_coord$} is formulated similarly to that in the local stage:
\begin{equation}
  \mathcal{L}_{coord} = L1(P_{coord}, Y_{coord})
\end{equation}
The predicted nodes include not only positional information but also the probability that the node matches the ground-truth, with the true probability of the matching node being 1. The probability loss \replaced[id=R2, comment={A12}]{$\mathcal{L}_{prob}$}{$L_prob$} is expressed as follows:
\begin{equation}
  \mathcal{L}_{prob} = L1(P_{prob}, Y_{prob})
\end{equation}

\subsubsection{Reconstruction Loss Function}
\added[id=R1, comment={1}]{Reconstruction loss follows the naming convention used in DN-DETR and DINO for the denoising component, with subtle differences between the global and local stages. In the global stage, since nodes contain three types of information: coordinates, category, and direction, the reconstruction loss is defined as the weighted sum of node position loss, node direction loss, and node category loss, with its loss function construction and weight coefficients remaining the same as those in the global stage. In the local stage, where nodes contain only coordinate and category information, the reconstruction loss is the weighted sum of node position loss and node category loss, and its loss function construction and weight coefficients are the same as those of the losses in the local stage. Importantly, the key distinction between reconstruction loss and other similar losses in both stages lies in the ground truth assignment: reconstruction loss is computed using known ground truth values without requiring Hungarian matching, whereas losses in both stages rely on Hungarian matching to determine their ground truth assignments.}

\section{Experimental settings}
\label{sec:section4}
\subsection{Experimental datasets}
To validate the effectiveness of the proposed method, we conducted experiments on two publicly available datasets: City-Scale\added[id=R2, comment={A13}]{\cite{he2020sat2graph}} and SpaceNet\added[id=R2, comment={A13}]{\cite{van2018spacenet}}. The following is a detailed introduction to both datasets.
\subsubsection{City-Scale dataset}
The dataset\added[id=R2, comment={A13}]{\cite{he2020sat2graph}} comprises 180 RGB images, each with a resolution of $2048 \times 2048$ pixels and a spatial resolution of 1 meter per pixel. It covers 20 urban areas in the United States and was constructed specifically for road network extraction tasks. The annotation data come from OpenStreetMap. In our experiments, we followed the dataset splitting protocol from Sat2Graph\replaced[id=R2, comment={A13}]{\cite{he2020sat2graph}}{[11]}, dividing the dataset into 144 training images, 9 validation images, and 27 test images. For ease of training and inference, the images were cropped into $512 \times 512$ image tiles, with 128-pixel overlaps between adjacent tiles.
\subsubsection{SpaceNet3 dataset}
The dataset\added[id=R2, comment={A13}]{\cite{van2018spacenet}} was released as part of the SpaceNet challenge. The dataset contains 2549 remote sensing images, each with a resolution of $400 \times 400$ pixels and a spatial resolution of 1 meter per pixel. For training, validation, and testing purposes, the dataset was divided into 2040, 127, and 382 images, respectively.

\subsection{Implementation details}
\subsubsection{Data augmentation and experimental setup}
To improve the robustness of the model, random brightness, random contrast, and multiscale training data augmentation methods were applied during the training process. No data augmentation schemes were used during inference. The model was implemented via the PyTorch framework and trained on four NVIDIA RTX 3090 GPUs. The Adam with Weight Decay Fix (AdamW) optimizer was used; the initial learning rate was 0.0001, and it decayed to one-tenth of its value every 10 epochs. \added[id=R1, comment=10]{To ensure fairness in inference time, GLD-Road and all comparative methods were evaluated on a machine equipped with an Intel Xeon Gold 6148 CPU, 256 GB of memory, and a single NVIDIA RTX 3090 GPU.} In the experiments conducted on the City-Scale dataset\added[id=R2, comment={A13}]{\cite{he2020sat2graph}}, owing to the density of its urban road network, the number of query in the Query Extractor was set to 500 based on a statistical analysis. For the SpaceNet3 dataset\added[id=R2, comment={A13}]{\cite{van2018spacenet}}, the number of query was set to 300. In the Local Query Decoder, the number of queries was set to 8 for both datasets.
\subsubsection{Label process}
\added[id=R1, comment=9]{In the Cityscale and SpaceNet3 datasets, road networks are represented as undirected graphs using a dictionary structure, where each key corresponds to the coordinates of a road network node, and the values represent the coordinates of its adjacent nodes. The dataset is processed differently in the Global and Local stages.}

\added[id=R1, comment=9]{In the global stage, each dictionary key represents the coordinates of a road node, while the corresponding node direction is inferred from the relationships between the key and its adjacent nodes, as described in Section 3.3.1 on road node modeling representation. Specifically, as illustrated in Figure 3, each road node may have adjacent nodes in up to four directions. The direction labels are encoded as a 36-dimensional tensor, where indices 3, 9, 21, and 27 are assigned a value of 1, indicating the presence of roads in these directions, while all other indices remain 0, signifying the absence of roads.In the local stage, the labeling process follows a methodology similar to that of RNGDet and operates in two modes: the road segment mode and the road vertex mode. In road segments that do not contain intersections, local image patches of size \(128 \times 128\) are extracted at regular intervals of 20 pixels, along with the corresponding ground-truth raster maps. In the road vertex mode, unexplored road segments are first identified, after which the next node is selected to enter road segment mode for further labeling.}
% In both the Cityscale and SpaceNet3 datasets, road networks are represented using a dictionary data structure, where the keys denote road network nodes, and the values contain all adjacent nodes, forming an undirected graph. The datasets differ across the Global and Local stages, which we describe as follows:
% In the Global Stage, each road node's (x,y) coordinates serve as the key in the dictionary, while the direction of the node is determined based on the key's corresponding value, as outlined in Section 3.3.1 on road node modeling representation. Specifically, as illustrated in Figure 3, the node direction tensor is a 36-dimensional vector where indices 3, 9, 21, and 27 are set to 1, indicating the presence of roads in those directions, while all other indices remain 0, signifying the absence of roads.
% In the Local Stage, the labeling process is similar to that of RNGDet and follows two modes: the road segment mode and the road vertex mode. In road segments without intersections, local image patches of size 128x128 and ground-truth raster maps are sampled at intervals of 20 pixels. In the road vertex mode, unexplored road segments are identified, and the next node is selected to transition into road segment mode for further labeling.
\subsubsection{Training process}
The training process was conducted in two stages: Global and Local. In the global stage, ImageNet-pretrained weights were loaded, and the model was trained for 100 epochs. In the local stage, the model weights that yielded the best performance on the \replaced[id=R2, comment=1]{validation }{test} set were selected as the initial weights for the RGB Backbone and Query Extractor modules to accelerate the model convergence process; thus, only 10 epochs of training were required for the local stage.
\subsubsection{Inference process}
On the City-Scale dataset, owing to its large image size, the global stage used a $512 \times 512$ sliding window for inference, with an overlap of 128 pixels. In the local stage, the number of retrieval steps was limited to maximum of 5. On the SpaceNet3 dataset, the global stage used full-image inference, and the local stage similarly limited the number of retrieval steps to a maximum of 5 to reduce the accumulated error.Based on the configuration of RNGDet\cite{xu2022rngdet} and RNGDet++\cite{xu2023rngdet++}, in the local stage on both datasets, a $128 \times 128$ patch centered on the road endpoints was cropped as the model input.

\subsection{Evaluation metrics}
The focus of the current GLD-Road research is to improve the topological integrity of road network structures. The existing methods for evaluating the accuracy of road network topologies involve two main aspects: local topological connection accuracy and global topological connection accuracy. The commonly used metrics include TOPO\added[id=R2, comment={A13}]{\cite{biagioni2012inferring}} and APLS\added[id=R2, comment={A13}]{\cite{van2018spacenet}}.
\subsubsection{TOPO method}
This method\added[id=R2, comment={A13}]{\cite{biagioni2012inferring}} evaluates the local topological similarity between the ground-truth map and the predicted map. First, seed points are selected from the ground-truth map, and corresponding points are searched in the predicted map based on the matching conditions set for the angles and positions around these seed points. If a matching point is found, the associated seed point is marked as a "valid seed point." For each valid seed point, all nodes within a certain threshold range are traversed in both the ground-truth and the predicted maps, enabling the extraction two corresponding subgraphs. By calculating the proportion of seed points that satisfy the matching conditions, the similarity between the two subgraphs can be assessed, ultimately resulting in average precision, recall, and F1 score values for all the sampled points.
\subsubsection{APLS method}
APLS\added[id=R2, comment={A13}]{\cite{van2018spacenet}} can be employed to assess the overall similarity between the predicted and ground-truth maps by focusing on differences among the shortest paths between pairs of vertices within a graph. The process begins by randomly selecting a subset of vertices from the ground-truth map and identifying their corresponding matches in the predicted map. The global topological structure difference between the two graphs is quantified by calculating the total variation in the shortest path distances between the matching vertex pairs in the ground-truth and predicted maps.
\begin{align}
  &S_{P \rightarrow T} = \nonumber \\
  &1 - \frac{1}{M} \sum_{(v_1, v_2) \in V} \min \left( 1, \frac{\left| L(v_1, v_2) - L(\hat{v}_1, \hat{v}_2) \right|}{L(v_1, v_2)} \right)
\end{align}
% \begin{align}
%   &\text{MSDA}\left( q, p_q, \{ x^l \}_{l=1}^{L} \right) = \nonumber \\
%   &\sum_{m=1}^{M} W_m \left[ \sum_{l=1}^{L} \sum_{k=1}^{K} A_{mlqk} \cdot W_m' x^l \left( \phi_l \left( p_q \right) + \Delta p_{mlqk} \right) \right]
% \end{align}
$V$ represents the set of sampled vertex pairs, and $M$ represents the total number of samples. The APLS metric is defined as follows:
\begin{equation}
APLS = \frac{S_{P \rightarrow T} S_{T \rightarrow P}}{S_{P \rightarrow T} + S_{T \rightarrow P}}
\end{equation}

\subsection{Comparison methods}
In the experimental comparison, we compared GLD-Road with five other methods. To evaluate the TOPO and APLS metrics, all the results are represented in the form of $G=(V, E)$. The following is a brief introduction to the comparison methods.

\textbf{DeepRoadMapper}\added[id=R2, comment={A17}]{\cite{mattyus2017deeproadmapper}}: This method relies on iterative tracking, beginning with the initialization of road pixels derived from the output of a segmentation network. It then reconnects any broken road segments by applying a shortest-path search algorithm.

\textbf{RoadTracer}\added[id=R2, comment={A17}]{\cite{bastani2018roadtracer}}: This is an iterative tracking-based road extraction method that uses a CNN-based decision function to guide an iterative search process, gradually retrieving and constructing a road network graph.

\textbf{Sat2Graph}\added[id=R2, comment={A17}]{\cite{he2020sat2graph}}: This is an end-to-end graph-based method that encodes the given road network into a high-dimensional tensor. The results are predicted by a deep network, and the road nodes are connected through postprocessing steps to generate a complete road network.

\textbf{RNGDet}\added[id=R2, comment={A17}]{\cite{xu2022rngdet}}: This is an end-to-end road extraction method based on DETR that uses an iterative tracking strategy to generate a road network structure.

\textbf{RNGDet++}\added[id=R2, comment={A17}]{\cite{xu2023rngdet++}}: This is an improved version of RNGDet that further incorporates a multiscale feature fusion module to achieve enhanced detection performance.

\textbf{IS-RoadDet}\cite{10720904}: This is a method that represents the road network as road segment instances and road endpoints.

\textbf{SamRoad}\cite{10678570}: This is an end-to-end method that uses SAM\cite{kirillov2023segment} as a feature extractor and connects adjacent road nodes based on their features.

\section{Experimental results and discuss}
\label{sec:section5}

\begin{figure*}[!ht]
  \centering
  \includegraphics[width=0.9\textwidth]{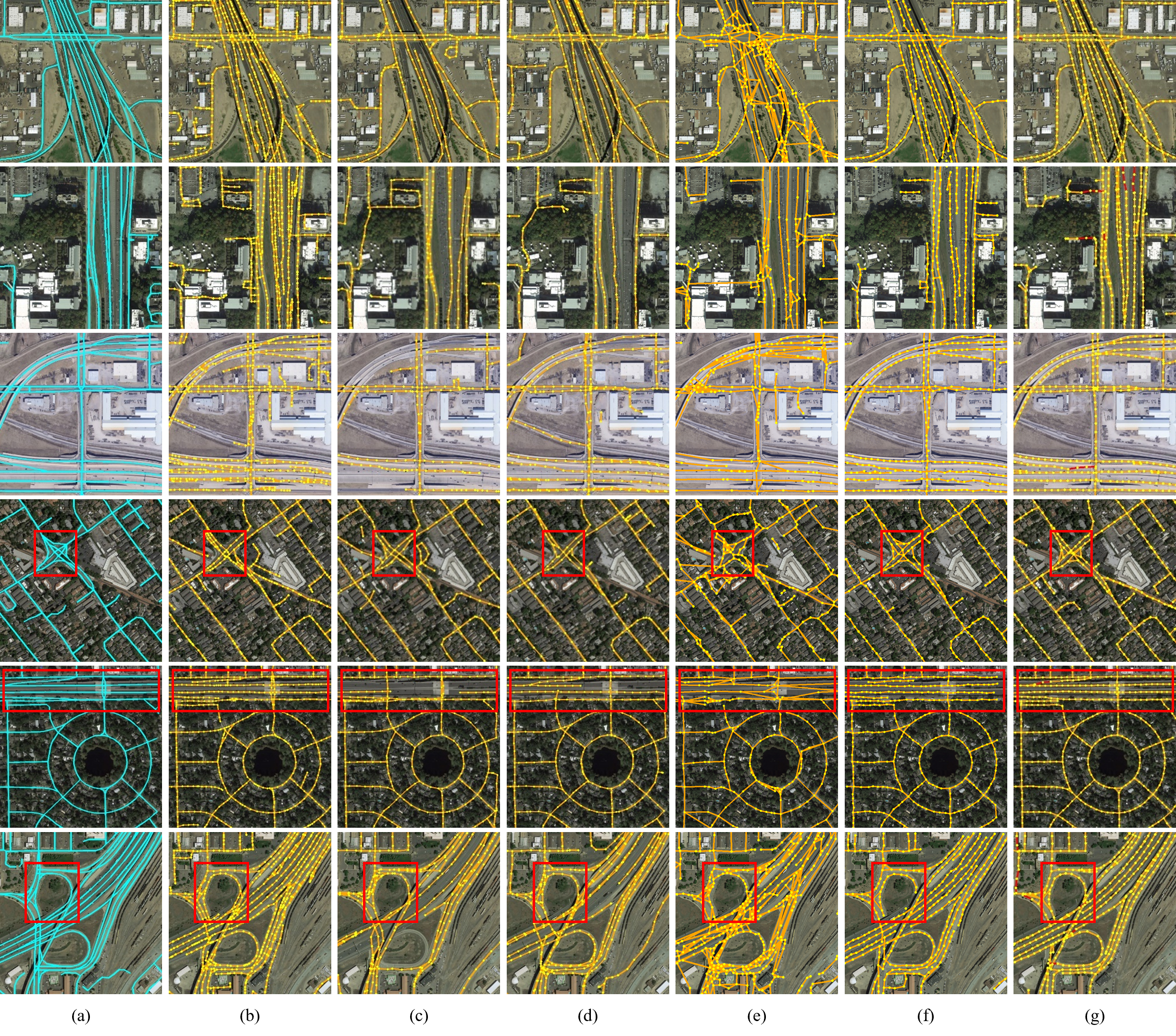}
\caption{(a) Ground-truth, (b) Sat2Graph(ECCV2020), (c) RNGDet(TGRS2022), (d) RNGDet(RAL2023), (e)IS-RoadDet(TGRS2025), (f)SamRoad(CVPRW2024), and (g) GLD-Road. Comparison among the visualized results produced for a portion of the City-Scale dataset. The cyan lines represent the ground-truth, the orange lines represent the predicted road network, and the yellow dots represent the nodes. Notably, the red line segments in (e) indicate the iterative retrieval results derived from the Local Query Decoder.}\label{fig:cs20_r}
\end{figure*}

\subsection{Experiments conducted on the City-Scale dataset}
On the City-Scale dataset, a quantitative analysis of the proposed GLD-Road approach and several existing comparison methods is provided in terms of the TOPO and APLS connectivity metrics as well as the inference times of the various methods, as shown in \autoref{tab:cs_r_tab}.
The data in \autoref{tab:cs_r_tab} indicate that the GLD-Road method outperformed the other comparison methods in terms of two topological accuracy metrics: TOPO-F1, and APLS. Specifically, its TOPO-F1, and APLS values were  1.05\%, and 1.9\% higher, respectively, than those of the best comparison method. Although GLD-Road does not achieve the best performance in either the TOPO-P or TOPO-R metric, it attains a better balance between precision and recall across the TOPO metrics. In terms of inference time, the GLD-Road method exhibited high efficiency. If only the Global Query Decoder module was used to generate the initial road network structure, the inference time was 0.11 hours, demonstrating higher inference efficiency than all other comparison methods. Even at this stage, the APLS and TOPO-F1 metrics of GLD-Road were already superior to those of all the other methods. A further analysis of the last two rows in \autoref{tab:cs_r_tab} reveals that after introducing the Local Query Decoder, the APLS accuracy of GLD-Road increased by an additional 1.13 percentage points, reaching 69.66\%, whereas the inference time remained at only 0.38 hours, which was still faster than those of the majority of the other methods. This finding indicates that while achieving higher accuracy, GLD-Road can still maintain an efficient inference speed.

\begin{table}[ht]
  \centering
\caption{Quantitative results obtained on the City-Scale dataset. All TOPO and APLS metrics are in percentage. The best results are highlighted in bold.}
\resizebox{\columnwidth}{!}{
\begin{tabular}{lccccc}
\hline
\textbf{Method} & \textbf{TOPO-P} & \textbf{TOPO-R} & \textbf{TOPO-F1} & \textbf{APLS} & \textbf{Infer. Time} \\
\hline
DeepRoadMapper & 73.57 & 76.61 & 75.05 & 53.18 & 2.71h \\
RoadTracer & 74.41 & 58.68 & 65.62 & 58.89 & 1.13h \\
Sat2Graph & 80.70 & 72.28 & 76.26 & 63.14 & 0.64 h \\
RNGDet & 85.97 & 69.78 & 76.87 & 65.75 & 2.93 h \\
RNGDet++ & 85.65 & 72.58 & 78.44 & 67.76 & 4.82 h \\
IS-RoadDet & 68.97 & \textbf{79.75} & 73.76 & 65.65 & 6.75 h \\
SamRoad & \textbf{90.05} & 67.71 & 77.09 & 66.96 & 0.19 h \\
GLD-Road (Global) & 84.98 & 74.94 & \textbf{79.55} & 68.53 & \textbf{0.11 h} \\
GLD-Road & 83.81 & 75.77 & 79.49 & \textbf{69.66} & 0.38 h \\
\hline
\end{tabular}%
}\label{tab:cs_r_tab}
\end{table}

To more intuitively demonstrate the effectiveness of the GLD-Road method, \autoref{fig:cs20_r} presents a comparison among the visual results produced by GLD-Road and the other highly accurate comparison methods on the City-Scale dataset; these methods included Sat2Graph, RNGDet, RNGDet++, SamRoad, and IS-RoadDet. The columns in \autoref{fig:cs20_r} correspond to the detection results produced by different methods for a single scenario, while the rows show the performance attained by each method in different scenarios. The first two rows of \autoref{fig:cs20_r} illustrate that while Sat2Graph and SamRoad were able to extract most roads, obvious disconnections were presented in the road network, which severely impacted the overall connectivity level. In contrast, RNGDet and RNGDet++ performed better in terms of connectivity but missed some roads, weakening their overall road network extraction effects. \added[id=R2, comment={3,R3:3}]{IS-RoadDet, on the other hand, produced a large number of incorrect connections.} Compared with these methods, GLD-Road had a higher road recall rate with fewer disconnections, making its detection results closer to the ground-truth. In the scenario shown in row 3 of \autoref{fig:cs20_r}, the first four methods exhibit issues such as disconnections, missing roads, and chaotic connections. \added[id=R2, comment={3,R3:3}]{SamRoad achieves good performance in road network detection; however, the roads detected by SamRoad appear overly curved, which does not align with the typically straight nature of actual roads.} In contrast, GLD-Road produces more accurate and visually coherent detection results compared with the other methods. Rows 4 and 5 of \autoref{fig:cs20_r} show the results obtained for scenarios with complex intersections: Sat2Graph, RNGDet, RNGDet++, and IS-RoadDet all exhibited varying degrees of incorrect connections or missing connections. In contrast, GLD-Road and SamRoad performed more accurately in terms of handling intersecting roads, and its detection results were highly consistent with the ground-truth labels. Rows 5 and 6 display the results obtained for the ring road scenarios. In row 5, Sat2Graph produced multiple disconnections in the ring roads, and in row 6, its results reveal confusion at the ring road connections. RNGDet and RNGDet++ both exhibited missed or incorrect connections in the ring road scenarios shown in row 6. \added[id=R2, comment={3,R3:3}]{Although IS-Road connects all detected roads, it introduces a large number of incorrect connections. In the red box of the visualization result in row 5, SamRoad shows an isolated branch road that is not connected to the main road. Similarly, in the red box of row 6, the detected roundabout is also isolated.} In comparison, GLD-Road produced more accurate and clear detection results in the ring road scenarios, demonstrating superior performance to that of the other methods.

From the visual results obtained in these specific regions, it is evident that GLD-Road consistently delivered better detection results across various scenarios, particularly in long straight road and ring road scenarios.

\begin{figure*}[!ht]
  \centering
  \includegraphics[width=0.9\textwidth]{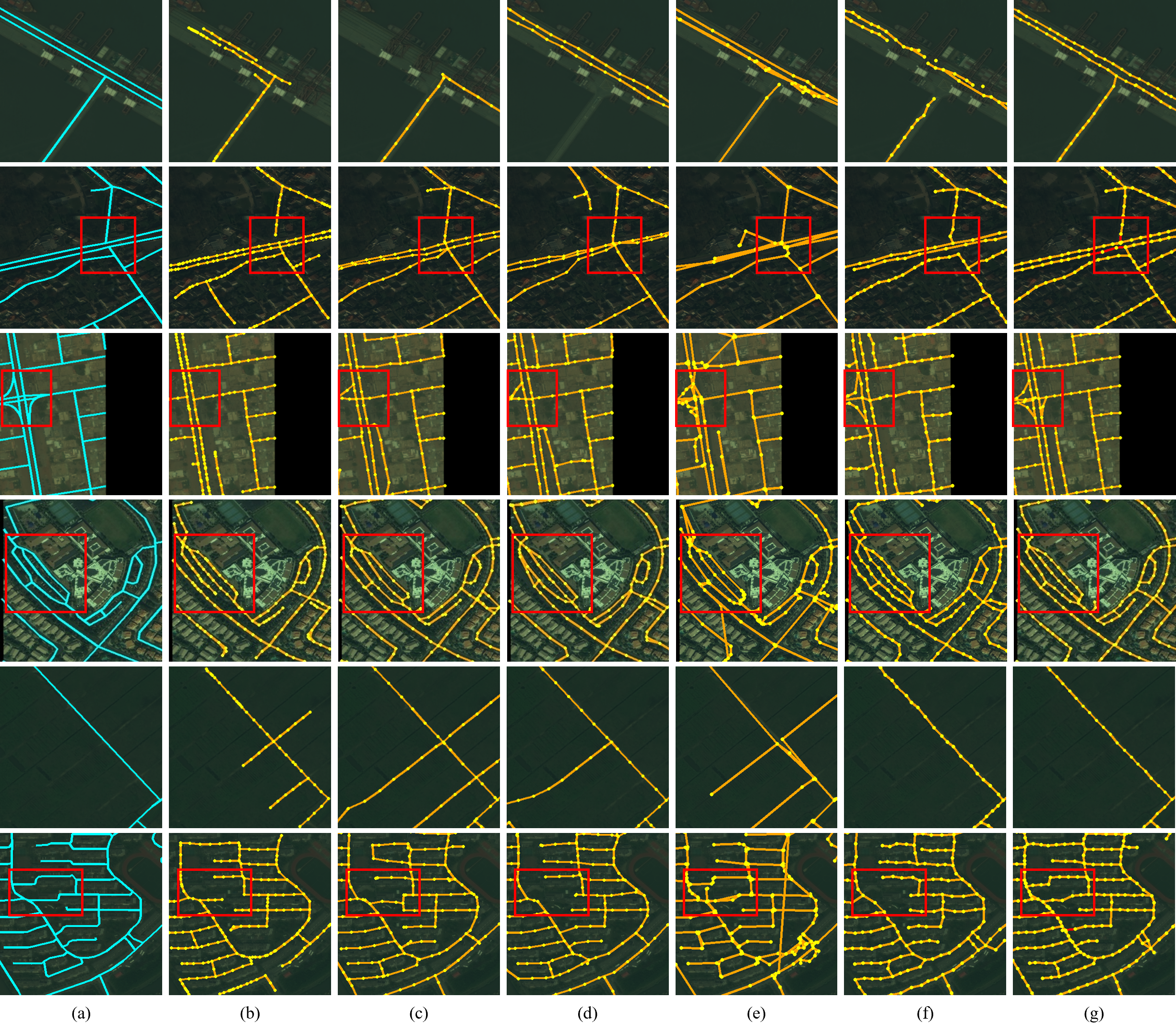}
\caption{ (a) Ground-truth, (b) Sat2Graph(ECCV2020), (c) RNGDet(TGRS2022), (d) RNGDet(RAL2023), (e)IS-RoadDet(TGRS2025), (f)SamRoad(CVPRW2024), and (g) GLD-Road. Comparison among the visual results produced for a portion of the SpaceNet3 dataset. The cyan lines represent the ground-truth, the orange lines represent the predicted road network, and the yellow dots represent the nodes. Notably, the red line segments in (e) indicate the iterative retrieval results derived from the Local Query Decoder.}\label{fig:sn3_r}
\end{figure*}

\subsection{Experiments conducted on the SpaceNet3 dataset}
\autoref{tab:sn3_r_tab} presents the quantitative comparison results produced by GLD-Road and other methods on the SpaceNet3 dataset. As shown in \autoref{tab:sn3_r_tab}, GLD-Road outperformed the other methods in terms of the TOPO-R, TOPO-F1, and APLS metrics, exceeding the second-best comparison method by 1.6\%, 2.21\%, and 0.67\%, respectively. Additionally, the inference time of GLD-Road was faster than that of the majority of the other comparison methods, demonstrating higher inference efficiency while maintaining high accuracy. The last two columns also show that the TOPO-F1 and APLS accuracies achieved during the Local Query Decoder stage improved by 0.21\% and 0.58\%, respectively, whereas the inference time increased by only 0.05 hours, which was much lower than the 0.28 hours required for the City-Scale dataset. This difference was due mainly to the smaller image areas, simpler road network structures, and fewer road endpoints contained in the SpaceNet3 dataset. Overall, the results presented in \autoref{tab:sn3_r_tab} demonstrate that GLD-Road not only achieved higher road network topological accuracy on the SpaceNet3 dataset but also exhibited faster inference efficiency. Compared with the other existing methods, GLD-Road performed the best in terms of TOPO-R, TOPO-F1, and APLS.

\begin{table}[ht]
\centering
\caption{Quantitative results obtained on the SpaceNet3 dataset. All TOPO and APLS metrics are in percentage. The best results are highlighted in bold.}
\resizebox{\columnwidth}{!}{
\begin{tabular}{lccccc}
\hline
\textbf{Method} & \textbf{TOPO-P} & \textbf{TOPO-R} & \textbf{TOPO-F1} & \textbf{APLS} & \textbf{Infer. Time} \\
\hline
DeepRoadMapper & 81.44 & 73.14 & 77.07 & 61.92 & 1.79h \\
RoadTracer & 77.48 & 63.51 & 69.8 & 57.84 & 0.94h \\
Sat2Graph & 85.93 & 76.55 & 80.97 & 64.43 & 0.52 h \\
RNGDet & 90.91 & 73.25 & 81.13 & 65.61 & 1.68 h \\
RNGDet++ & 91.34 & 75.24 & 82.51 & 67.73 & 2.75 h \\
IS-RoadDet & 87.44 & 51.51 & 64.83 & 53.52 & \textbf{0.11 h} \\
SamRoad & 83.54 & 75.27 & 79.19 & 71.14 & 0.29 h \\
GLD-Road (Global) & \textbf{93.16} & 77.34 & 84.51 & 71.23 & 0.31 h \\
GLD-Road & 92.51 & \textbf{78.15} & \textbf{84.72} & \textbf{71.81} & 0.36 h \\
\hline
\end{tabular}
}\label{tab:sn3_r_tab}
\end{table}

To provide a more intuitive comparison among the performances of different methods, \autoref{fig:sn3_r} presents the visual results produced by various methods over specific areas. In the first row, Sat2Graph yielded dense road points in the long straight road regions, with a fragmented predicted road network. In contrast, both RNGDet and RNGDet++ produced omissions in the same region, leading to a decrease in their overall road network recall rates. \added[id=R2, comment={3,R3:3}]{IS-Road exhibited duplicated road predictions, while SamRoad produced multiple isolated road segments.} The results produced by GLD-Road, however, demonstrated better connectivity and recall. In the complex intersections highlighted by the red boxes in the second and third rows, the other methods showed significant discrepancies relative to the ground-truth, whereas GLD-Road provided more complete detection results. In the fourth row, the results derived from Sat2Graph contain missing sections of connected roads, and while RNGDet and RNGDet++ detected these road structures, incorrect connections affected their overall topological accuracy. In comparison, GLD-Road, IS-RoadDet and SamRoad not only achieved complete road network detection but also ensured the correctness of the topological structure. The fifth row shows that the first four methods extracted many incorrect roads within the network. Some of the roads extracted by SamRoad appear curved, which does not conform to the typically straight nature of real roads, whereas the detection results of GLD-Road are almost perfectly aligned with the ground-truth.
\added[id=R2, comment={3,R3:3}]{In the areas of the last row in red boxes, none of the other comparison methods—except for GLD-Road and IS-RoadDet—were able to fully detect the road network. However, IS-RoadDet produced many incorrect connections in other areas, such as the bottom-right region. GLD-Road, by contrast, achieved superior local connectivity.}

\subsection{Ablation study}
To quantitatively analyze and verify the rationality of each module contained in GLD-Road, ablation experiments were conducted on the City-Scale dataset.
\subsubsection{Impact of the iterative step number}
\begin{figure}[!htbp]
  \centering
  \includegraphics[width=\columnwidth]{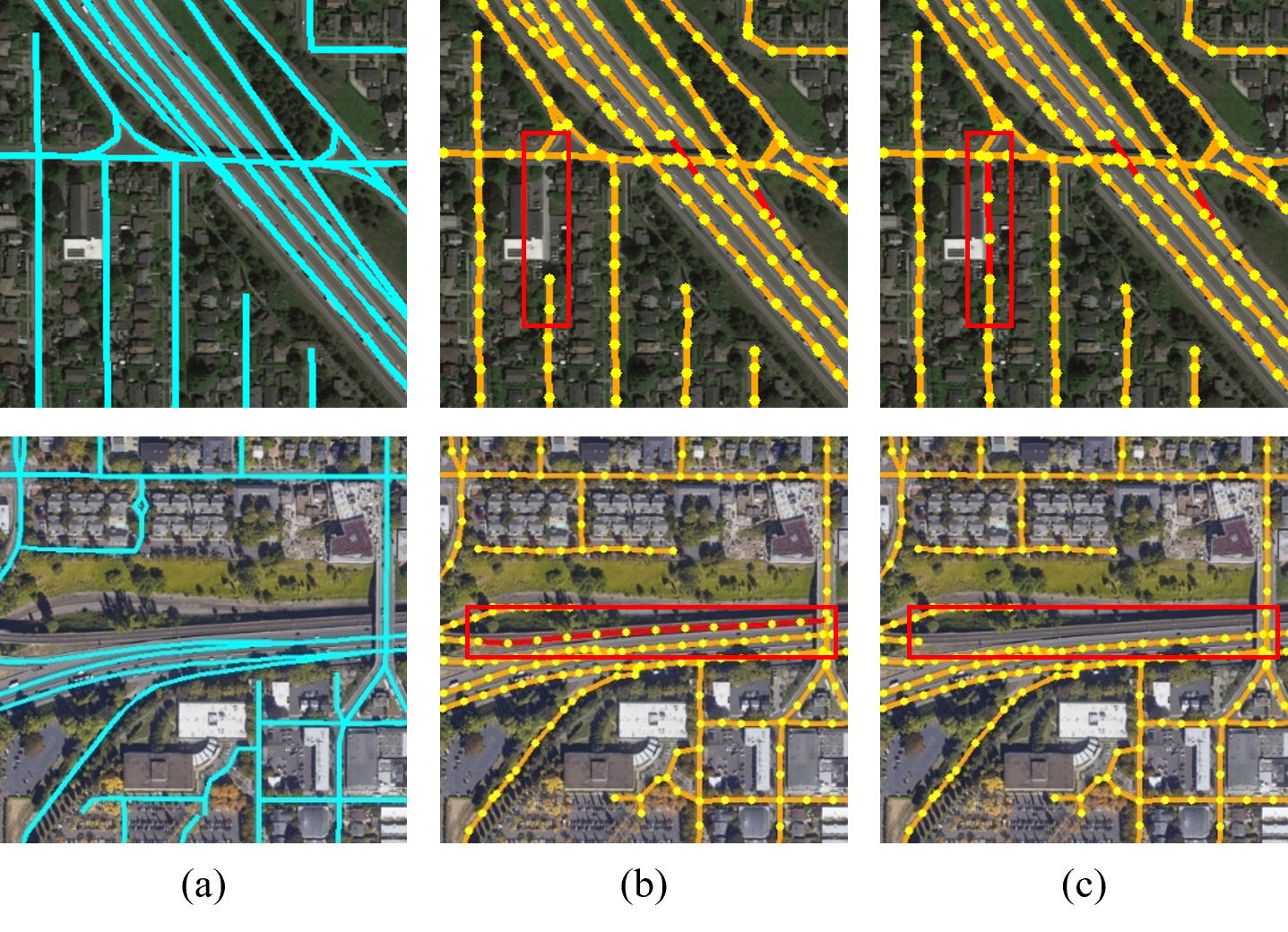}
\caption{(a) Ground-truth, (b) Results obtained with 1 and 20 steps, and (c) Connection results obtained with 5 steps. The cyan lines represent the ground-truth, the orange lines represent the predicted road network, the yellow dots represent the nodes, and the red line segments indicate the iterative retrieval results derivedfrom the Local Query Decoder.}\label{fig:iter_num}
\end{figure}

We investigated the effects of different retrieval step lengths on the TOPO-F1, APLS, and retrieval time results. The quantitative comparison is shown in \autoref{tab:iter_num_tab}. First, continuously increasing the retrieval step length did not necessarily improve the connection accuracy of the road network. When the retrieval step length was 5, APLS reached their peak values at 69.66\%, respectively. \autoref{fig:iter_num} shows that when the retrieval step length was too short, the scenario in the red box in the first row of \autoref{fig:iter_num} occurred, where distant disconnected roads could not be connected. When the step length was too long, overprediction of the disconnected roads occurred, as shown in the second row of \autoref{fig:iter_num}. Although TOPO-F1 slightly decreased compared to the case with step = 1, APLS improved significantly when the step size was set to 5. Therefore, a step length of 5 served as a more favorable hyperparameter choice. Second, comparing the APLS accuracies between adjacent rows of data, the APLS accuracy improvement was greatest when the step length was 1 compared with the previous row, indicating that most disconnections in the initial road network were very short-distance disconnections. This conclusion is visually supported by the red line segment contained in the first row of \autoref{fig:iter_num}(c).

\begin{table}[ht]
\centering
\caption{Ablation study results obtained with different step lengths for the Local Query Decoder. TOPO-F1 and APLS metrics are in percentage. The best results are highlighted in bold.}
\resizebox{\columnwidth}{!}{
\begin{tabular}{cccc}
\hline
\textbf{Iter. num.} & \textbf{TOPO-F1 (\%)} & \textbf{APLS (\%)} & \textbf{Time} \\
\hline
0 & 79.55 & 68.53 & 0 s \\
1 & \textbf{79.52} & 69.01 & 574 s \\
5 & 79.49 & \textbf{69.66} & 1045 s \\
10 & 79.23 & 69.24 & 1386 s \\
20 & 78.99 & 68.47 & 1595 s \\
\hline
\end{tabular}
}\label{tab:iter_num_tab}
\end{table}

\subsubsection{Impact of adding the backbone and Local Query Decoder}
We also evaluated the effectiveness of the Local Query Decoder with different backbones. As shown in \autoref{tab:backbone_tab}, after adding the Local Query Decoder, significant accuracy improvements were observed across all six backbones, with an average APLS improvement of 1.06\% and an average TOPO-F1 improvement of 0.25\%. Additionally, it was evident that the performance improvements were greater in lower-performing baseline models (R50 and R101) when the Local Query Decoder was added. This is because these baselines tended to have higher frequencies of fragmented road networks, and the addition of the Local Query Decoder more effectively completed their road networks. An analysis of the data in \autoref{tab:backbone_tab} reveals that the use of R50 as the backbone and the addition of the Local Query Decoder resulted in TOPO-F1 and APLS accuracies that also achieved competitive performance. This finding indicates that our method could still provide satisfactory road network results even with a less complex network structure.

\added[id=R2, comment=5]{However, as shown in \autoref{tab:backbone_tab}, when replacing the backbone with architectures of larger parameter sizes, GLD-Road shows improvements in the TOPO-F1 metric, but the increase in the APLS metric is relatively small, with even a slight decline observed. The reason for this phenomenon is that simply using a larger backbone does not address the road connectivity issues in complex road scenarios. This observation suggests that further improvements in road network connectivity may not be achieved merely by adopting larger backbones. Instead, designing new modules or exploring new representations of road networks may prove more effective.}

\begin{table}[ht]
\centering
\caption{Ablation study results obtained regarding the effectiveness of different backbones and the addition of the Local Query Decoder. TOPO-F1 and APLS metrics are in percentage.}
\resizebox{\columnwidth}{!}{
\begin{tabular}{ccccc}
\hline
\textbf{Backbone} & \textbf{Local Query Decoder} & \textbf{TOPO-F1} & \textbf{APLS} & \textbf{Infer. Time} \\
\hline
R50 & × & 76.42 & 63.82 & 302s \\
R50 & \checkmark & \textbf{77.05} & \textbf{65.31} & 1485s \\
R101 & × & 76.92 & 64.61 & 342s \\
R101 & \checkmark & \textbf{78.31} & \textbf{66.15} & 1641s \\
Swin-Tiny & × & 79.21 & 68.37 & 321s \\
Swin-Tiny & \checkmark & \textbf{78.97} & \textbf{68.79} & 1492s \\
Swin-Small & × & \textbf{79.55} & 68.53 & 392s \\
Swin-Small & \checkmark & 79.49 & \textbf{69.66} & 1357s \\
Swin-Base & × & \textbf{79.74} & 68.72 & 465s \\
Swin-Base & \checkmark & 79.71 & \textbf{69.59} & 1682s \\
Swin-Large & × & 80.03 & 68.94 & 542s \\
Swin-Large & \checkmark & \textbf{80.11} & \textbf{69.88} & 2129s \\

% \textcolor{deepgreen}{Swin-Base} & \textcolor{deepgreen}{×} & \textcolor{deepgreen}{\textbf{79.74}} & \textcolor{deepgreen}{68.72} & \textcolor{deepgreen}{465s} \\
% \textcolor{deepgreen}{Swin-Base} & \textcolor{deepgreen}{\checkmark} & \textcolor{deepgreen}{79.71} & \textcolor{deepgreen}{\textbf{69.59}} & \textcolor{deepgreen}{1682s} \\
% \textcolor{deepgreen}{Swin-Large} & \textcolor{deepgreen}{×} & \textcolor{deepgreen}{80.03} & \textcolor{deepgreen}{68.94} & \textcolor{deepgreen}{542s} \\
% \textcolor{deepgreen}{Swin-Large} & \textcolor{deepgreen}{\checkmark} & \textcolor{deepgreen}{\textbf{80.11}} & \textcolor{deepgreen}{\textbf{69.88}} & \textcolor{deepgreen}{2129s} \\
\hline
\end{tabular}
}
\label{tab:backbone_tab}
\end{table}

\subsubsection{Impacts of different directional descriptor dimensions}
To verify the effectiveness of the proposed directional descriptor, we conducted a systematic ablation study on the dimensions of the directional descriptor. Specifically, we selected seven different sets of hyperparameters, using intervals of $\frac{\pi}{6}$, $\frac{\pi}{8}$, $\frac{\pi}{12}$, $\frac{\pi}{18}$, $\frac{\pi}{36}$, $\frac{\pi}{72}$, and $\frac{\pi}{360}$ as the angles between adjacent directions to explore the impact of this parameter on the resulting road network extraction accuracy. Since the directional descriptor only affects the global stage, we compare the performance solely based on the road network accuracy at the global stage. The data in \autoref{tab:direct_tab} indicate that as the angular interval between directions decreased, the accuracy improved to a certain extent. However, when the angle was further reduced to $\frac{\pi}{72}$ or $\frac{\pi}{360}$, the accuracy decreased. These results indicate that $\frac{\pi}{36}$ was the optimal interval angle for the directional descriptor, yielding the best road network detection results.

\added[id=R2, comment={A14}]{As an additional note, regarding the phenomenon where accuracy decreases with finer angular intervals, we believe two main factors contribute to this outcome. First, excessively fine direction discretization may introduce inconsistencies in similar scenarios, leading to ambiguities that hinder the model's convergence. Second, smaller intervals may cause the model to focus excessively on fine details, leading to overfitting and reducing its generalization ability on new data. The combined effect of these two factors results in a decline in accuracy.}

\begin{table}[ht]
  \centering
\caption{Ablation study results obtained with different directional descriptor dimensions.}
\resizebox{\columnwidth}{!}{
\begin{tabular}{ccc}
\hline
\textbf{Interval Angle} & \textbf{TOPO-F1 (\%)} & \textbf{APLS (\%)} \\
\hline
$\pi/6$ & 75.84 & 64.97 \\
$\pi/8$ & 75.1 & 64.62 \\
$\pi/12$ & 77.48 & 66.81 \\
$\pi/18$ & 79.15 & 67.99 \\
$\pi/36$ & \textbf{79.55} & \textbf{68.53} \\
$\pi/72$ & 78.67 & 67.93 \\
$\pi/360$ & 78.89 & 67.56 \\
\hline
\end{tabular}
}\label{tab:direct_tab}
\end{table}

\subsubsection{Impacts of Different Noise Scales $\lambda$ for Denoising}
As shown in \autoref{tab:denoising_tab}, different values of $\lambda$ had significant effects on the connection accuracy metrics. The highest accuracy was achieved when $\lambda$=10. When $\lambda$=1 or 5, the model was unable to effectively identify road nodes in densely populated areas, resulting in prediction confusion and connection errors, which led to lower APLS accuracy values. When $\lambda$=20, more missing nodes were observed at road intersections, which also caused the APLS accuracy to decrease compared with that achieved when $\lambda$=10. However, since TOPO metrics measure accuracy in local regions, the TOPO-P accuracy was higher when $\lambda$=1 due to the fewer predicted nodes, whereas the TOPO-R metric was lower. Taking everything into consideration, the TOPO-F1 and APLS metrics were both highest when $\lambda$=10.

\added[id=R2, comment={A15}]{In addition, the first row in \autoref{tab:denoising_tab} shows the metric results without applying the denoising training strategy, and the comparison clearly indicates the effectiveness of this strategy.}

\begin{table}[ht]
  \centering
\caption{Comparison among the accuracies achieved with different $\lambda$ values.}
\resizebox{\columnwidth}{!}{
\begin{tabular}{ccccc}
\hline
$\lambda$ & \textbf{TOPO-P (\%)} & \textbf{TOPO-R (\%)} & \textbf{TOPO-F1 (\%)} & \textbf{APLS (\%)} \\
\hline
no dn & 86.03 & 60.79 & 71.22 & 63.09 \\
1 & \textbf{86.32} & 67.99 & 76.06 & 64.93 \\
5 & 85.75 & 72.23 & 78.39 & 67.83 \\
10 & 83.81 & \textbf{75.77} & \textbf{79.49} & \textbf{69.66} \\
20 & 84.91 & 73.02 & 78.51 & 68.45 \\
\hline
\end{tabular}
}\label{tab:denoising_tab}
\end{table}

\subsubsection{Impacts of Input Feature Preprocessing on the Connect Module}
\added[id=R2, comment={A8}]{In this section, we investigated the impact of two factors on road network extraction accuracy: (1) whether to apply separate preprocessing to coordinate features and the 36-dimensional directional descriptors before concatenation, and (2) the dimensionality of features input to the Transformer blocks within the Connect Module. Additionally, since both of these factors only affect the lobal stage, we conduct the ablation comparison based solely on the road network accuracy at the global stage. As shown in \autoref{tab:precat_mlp_ablation}, the first three experiments demonstrate that when the feature dimension is set to 64, the model achieves the best performance on both TOPO-F1 and APLS metrics. This result suggests that lower dimensions may fail to capture sufficient semantic information of road nodes, while excessively high dimensions could introduce redundant noise that hinders model learning, ultimately leading to performance degradation.}

\added[id=R2, comment={A8}]{Furthermore, a comparison of the last four experiments reveals that introducing independent MLP layers to the coordinate and directional features prior to concatenation leads to a decline in overall accuracy. This may be attributed to the separate MLPs disrupting the inherent representation patterns of the two feature types, weakening their mutual correlations and making it more difficult for subsequent modules to effectively capture the structural connectivity among road nodes.}

% \begin{table}[htbp]
%   \centering
%   \caption{Ablation study on Pre-Cat MLP and node feature dimensions.}
%   \begin{tabular}{cccc}
%   \toprule
%   \textbf{Pre-Cat MLP} & \textbf{Node Feature Dimensions} & \textbf{TOPO-F1} & \textbf{APLS} \\
%   \midrule
%   ✗ & 32  & 79.31 & 67.95 \\
%   ✗ & 64  & \textbf{79.55} & \textbf{68.53} \\
%   ✗ & 128 & 78.94 & 68.21 \\
%   ✓ & 64  & 77.54 & 67.46 \\
%   ✓ & 128 & 78.63 & 67.62 \\
%   \bottomrule
%   \end{tabular}
%   \label{tab:precat_mlp_ablation}
% \end{table}

\begin{table}[ht]
  \centering
\caption{Ablation study on Pre-Cat MLP and node feature dimensions.}
\resizebox{\columnwidth}{!}{
\begin{tabular}{cccc}
\hline
\textbf{Pre-Cat MLP} & \textbf{Node Feature Dimensions} & \textbf{TOPO-F1} & \textbf{APLS} \\
\hline
× & 32  & 79.31 & 67.95 \\
× & 64  & \textbf{79.55} & \textbf{68.53} \\
× & 128 & 78.94 & 68.21 \\
\checkmark & 64  & 77.54 & 67.46 \\
\checkmark & 128 & 78.63 & 67.62 \\
\hline
\end{tabular}
}\label{tab:precat_mlp_ablation}
\end{table}

\subsection{Discuss}
\begin{figure}[!htbp]
  \centering
  \includegraphics[width=\columnwidth]{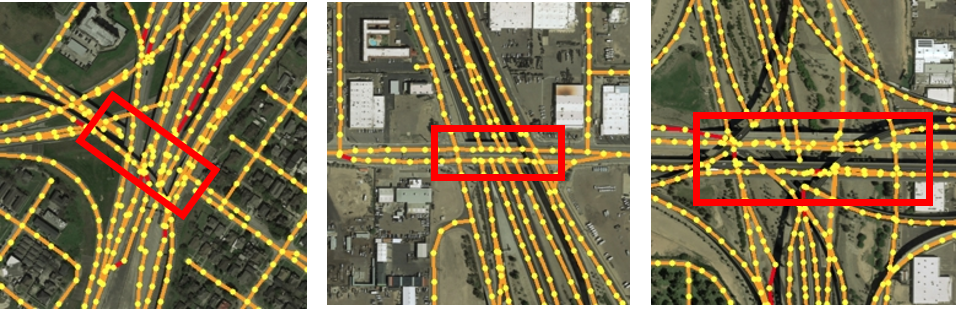}
\caption{Failed Cases of Road Network Extraction. Red boxes indicate road network disconnection areas. The orange lines represent the predicted road network, the yellow dots represent the nodes, and the red line segments indicate the iterative retrieval results derivedfrom the Local Query Decoder.}\label{fig:discuss}
\end{figure}

This section primarily discusses the limitations of the GLD-Road model and directions for future improvement. In terms of road network extraction performance, as highlighted by the red box area in \autoref{fig:discuss}, GLD-Road struggles to accurately extract complete road networks in overpass scenarios. This deficiency arises mainly from two factors: first, remote sensing images are two-dimensional and lack the capability to represent the height information of road networks, making it difficult to extract the full extent of overpass roads. To address this, future work could consider integrating remote sensing imagery with GPS information to construct a more comprehensive road network structure. Second, GLD-Road models roads by discretizing them into interconnected nodes. While this method avoids significant deviations or the loss of entire road segments, its limitation lies in the interaction only between adjacent nodes. When the predicted adjacent nodes are spaced far apart, the model may fail to connect them correctly or even result in disconnections, which is particularly evident in overpass scenarios. To tackle this issue, future exploration could focus on representations based on entire road segments and draw inspiration from the multi-point attention mechanism of StreamMapNet\cite{yuan2024streammapnet} to enhance the completeness of road detection and improve the model's convergence speed.

In terms of extraction efficiency, GLD-Road outperforms other comparative methods. However, the model requires two-stage training and is based on the DETR framework, which does not offer a significant advantage in training convergence speed compared to other methods. Although the local stage can utilize the weights from the global stage to accelerate convergence, training the GLD-Road model still takes approximately 72 hours on four NVIDIA RTX 3090 GPUs to achieve optimal results. With the development of prompt learning models like SAM\cite{kirillov2023segment}, OMG-Seg\cite{li2024omg} and ChatGPT\cite{brown2020language}, constructing a single-stage global-local road network extraction model may become a future direction. Future models could first extract a primary road network and then use a prompt encoder and image encoder to integrate the raster information of the primary road network with imagery, achieving local fine-grained completion of the road network.

\section{Conclusion}
\label{sec:section6}
To address the issues of fragmented road networks in global iterative methods and low retrieval efficiency in local iterative methods, we propose a global-local decoding-based two-stage remote sensing image road network extraction model—GLD-Road. The method first uses global information to rapidly extract an initial road network, then performs local iterative searches on the road endpoints of the initial network to form the final road network. The two-stage strategy enables GLD-Road to achieve both fast parallel processing speed and strong iterative connectivity. To avoid the difficulty of parameter tuning in post-processing algorithms, we use 36-dimensional directional descriptors and train a small graph neural network model to connect nodes. To address the issue of node confusion in complex scenes, we introduce a denoising training module, which improves road node detection accuracy. Experiments on two public datasets demonstrate that GLD-Road outperforms state-of-the-art methods in terms of TOPO method and APLS. \added[id=R3, comment={4}]{It is worth noting that the global query decoding stage retains the high efficiency of global parallel methods. In the local query decoding stage, since only limited supplementary detection is required for locally missing areas, GLD-Road reduces the scope of global iterative search. GLD-Road also achieves the highest road network extraction efficiency on both public datasets.} Ablation studies further validate the rationality of GLD-Road's module design and hyperparameter selection. In the future, we will focus on addressing the incomplete road network issue in overpass scenarios and the limitations of two-stage training, further researching solutions to the challenging problem of road network extraction from remote sensing images.

\section*{Declaration of competing interest}
The authors declare that they have no known competing financial interests or personal
relationships that could have appeared to influence the work reported in this paper.

\section*{Acknowledgments}
This research was funded by the Science and Disruptive Technology Program of AIRCAS under Grant number E3Z21102, the Civil Aerospace Technology Pre-research Project of China's 14th Five-Year Plan, and the National Key Research and Development Program of China under Grant number 2021YFB3900503.

\bibliographystyle{elsarticle-num} % 使用编号样式
% % \bibliographystyle{plainnat} % 使用编号样式

\bibliography{myinfer}

\end{document}